\pdfoutput=1
\documentclass[11pt]{article}

\usepackage[final]{acl}

\usepackage{times}
\usepackage{latexsym} 

\usepackage[T1]{fontenc}
\usepackage[utf8]{inputenc}

\usepackage{microtype}

\usepackage{inconsolata}

\usepackage{graphicx}

\usepackage{tcolorbox}
\tcbuselibrary{breakable}

\usepackage{booktabs}
\usepackage{multirow}
\usepackage{paralist}
\usepackage{subfigure}
\usepackage{wrapfig}
\usepackage{caption}
\usepackage{amsmath,amssymb}
\usepackage{colortbl}
\usepackage{arydshln}
\usepackage[dvipsnames]{xcolor}
\usepackage{cleveref}
\crefformat{section}{\S#2#1#3} % see manual of cleveref, section 8.2.1
\crefformat{subsection}{\S#2#1#3}
\crefformat{subsubsection}{\S#2#1#3}

\title{VisRet: Visualization Improves Knowledge-Intensive \\ Text-to-Image Retrieval}
\author{
Di Wu$^1$\thanks{\hspace{0.5mm}Equal contribution.}, Yixin Wan$^1$\footnotemark[1], Kai-Wei Chang$^1$  \\
$^1$University of California, Los Angeles \\
{\tt \{diwu,kwchang\}@cs.ucla.edu}
}

\begin{document}
\maketitle

\begin{abstract}

Text-to-image retrieval (T2I retrieval) remains challenging because cross-modal embeddings often behave as bags of concepts, underrepresenting structured visual relationships such as pose and viewpoint. We propose \textbf{Visualize-then-Retrieve (VisRet)}, a retrieval paradigm that mitigates this limitation of cross-modal similarity alignment. VisRet first projects textual queries into the image modality via T2I generation, then performs retrieval within the image modality to bypass the weaknesses of cross-modal retrievers in recognizing subtle visual-spatial features. Across four benchmarks (Visual-RAG, INQUIRE-Rerank, Microsoft COCO, and our new Visual-RAG-ME featuring multi-entity comparisons), VisRet substantially outperforms cross-modal similarity matching and baselines that recast T2I retrieval as text-to-text similarity matching, improving nDCG@30 by 0.125 on average with CLIP as the retriever and by 0.121 with E5-V. For downstream question answering, VisRet increases accuracy on Visual-RAG and Visual-RAG-ME by 3.8\% and 15.7\% in top-1 retrieval, and by 3.9\% and 11.1\% in top-10 retrieval. Ablation studies show compatibility with different T2I instruction LLMs, T2I generation models, and downstream LLMs. VisRet provides a simple yet effective perspective for advancing in text-image retrieval.  Our code and the new benchmark are publicly available at \url{https://github.com/xiaowu0162/Visualize-then-Retrieve}.

\end{abstract}

\section{Introduction}

Text-to-Image (T2I) retrieval selects the most relevant images from a visual corpus given a textual query. It plays a crucial role in knowledge-intensive applications that augment textual inputs with rich visual evidence \citep{DBLP:conf/emnlp/ChenHCVC22, DBLP:conf/acl/Wang0LLCJHXG23, DBLP:conf/iclr/SheyninAPSGNT23, DBLP:journals/corr/abs-2412-10510}.

\begin{figure*}[ht!]
 \centering
 \vspace{-5mm}
 \includegraphics[width=\linewidth]{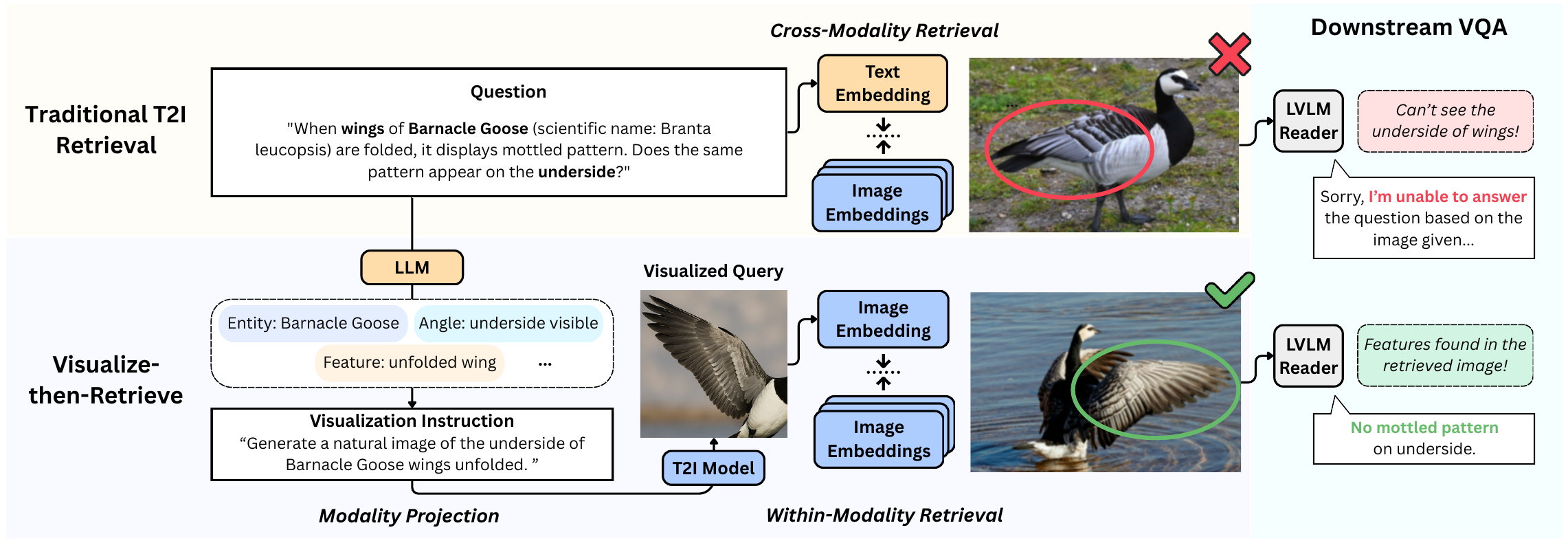}
 % \vspace{-2mm}
 \caption{An overview of VisRet. Compared to the traditional T2I retrieval pipeline, VisRet first projects the text query into the image modality via T2I generation and then performs within-modality retrieval.}
 \label{fig:main-figure}
\end{figure*}

A common approach embeds both the text query and image candidates into a shared representation space and ranks candidates by similarity \citep{frome2013devise, DBLP:journals/corr/KirosSZ14}. However, producing rankings that reflect fine-grained alignments between text and image remains a long-standing challenge. Prior studies show that cross-modal embeddings often behave like ``bags of concepts'' for similarity matching and fail to capture structured relationships among visual elements \citep{DBLP:conf/iclr/Yuksekgonul0KJ023, DBLP:conf/emnlp/KamathHC23}. For example, \Cref{fig:main-figure} presents a query that requires images of an entity (a Barnacle Goose) in a specific posture (wings unfolded). While the embedding model matches the entity type, it struggles to recognize subtler visual-spatial features such as the wing pose and the camera perspective (an upward-facing shot). To address these limitations, existing work improves embedding quality \citep{DBLP:conf/icml/RadfordKHRGASAM21, DBLP:journals/tmlr/YuWVYSW22}, performs query reformulation \citep{levy2023chatting}, or applies multi-stage reranking pipelines \citep{DBLP:journals/tmlr/0002STG24, DBLP:conf/aaai/0001B00KZGL25}. Nevertheless, these strategies remain constrained by the intrinsic difficulty of cross-modal similarity alignment because they cannot bypass the text-to-image similarity search stage.

We propose \textbf{Visualize-then-Retrieve (VisRet)}, a retrieval paradigm that decomposes T2I retrieval into two stages: text-to-image \textit{modality projection} followed by \textit{within-modality retrieva}l. VisRet first uses a text-to-image generation (T2I generation) model to visualize the textual query, with the goal of making key visual-spatial requirements explicit. The resulting visualization is then used as a query for direct image-to-image retrieval.

Compared to prior methods, VisRet offers two advantages. First, visualizations provide a more expressive and intuitive medium for representing compositional concepts such as entities, poses, and spatial relations. These requirements can be difficult to communicate through text alone, and encoding them exhaustively in a text query can hurt cross-modal matching due to limitations in embedding quality. In contrast, as shown in \Cref{fig:main-figure}, the visualized query can depict the desired entity, posture, and camera angle at the same time. Second, by operating entirely within the image modality during retrieval, VisRet avoids the weaknesses of cross-modal retrievers and leverages the stronger capacity of these retrievers in unimodal retrieval \citep{DBLP:journals/corr/abs-2502-03566}.

To evaluate VisRet, we conduct experiments with two widely used cross-modal embedding models across four benchmarks: Visual-RAG \citep{visual-rag}, INQUIRE-Rerank \citep{DBLP:conf/nips/VendrowPSBJABH24}, Microsoft COCO \citep{mscoco}, and Visual-RAG-ME, a new benchmark we introduce that features challenging visual feature comparison questions across multiple entities. VisRet substantially outperforms baseline cross-modal similarity matching and three strong baseline methods that cast T2I retrieval as text-to-text similarity matching (\cref{section-results-retrieval}). When CLIP \citep{DBLP:conf/icml/RadfordKHRGASAM21} is used as the retriever, VisRet improves nDCG@30 by 0.125 on average across the four benchmarks. With E5-V \citep{DBLP:journals/corr/abs-2407-12580} as the retriever, the improvement is 0.121. VisRet also improves downstream performance in retrieval-augmented generation (RAG) settings (\cref{section-results-qa}), increasing T2I question answering accuracy on Visual-RAG and Visual-RAG-ME by 3.8\% and 15.7\% in top-1 retrieval and by 3.9\% and 11.1\% in top-10 retrieval. Ablation studies and analyses show that VisRet is robust to the choice of instruction-generation LLM, retriever, and downstream reader LVLM, while performance depends more strongly on the T2I generation model, with stronger generators yielding larger gains. We further analyze repeated runs and find that aggregating multiple visualizations is beneficial, while using a single visualization only slightly reduces overall performance. Finally, although visualized queries are effective for retrieval, they are generally insufficient as standalone knowledge to answer downstream questions. Our code and the new Visual-RAG-ME benchmark will be released publicly to facilitate future research.

\section{Related Work}

\paragraph{T2I Retrieval Benchmarks}
Early T2I retrieval benchmarks evaluate whether a system can retrieve an image given its paired human-written caption. These datasets span multiple domains and include widely used benchmarks such as COCO \citep{mscoco}, Flickr8K \citep{DBLP:journals/jair/HodoshYH13}, Flickr30K \citep{DBLP:journals/tacl/YoungLHH14}, and Fashion200K \citep{han2017automatic}. As multimodal embedding models have matured, newer benchmarks have been introduced to assess retrieval in knowledge-intensive settings. Datasets such as WebQA \citep{DBLP:conf/cvpr/ChangCNGSB22}, INQUIRE \citep{DBLP:conf/nips/VendrowPSBJABH24}, Visual-RAG \citep{visual-rag}, and MRAG-Bench \citep{DBLP:journals/corr/abs-2410-08182} shift the focus from caption matching to retrieving images that provide the information needed to answer knowledge-intensive questions. These tasks evaluate T2I retrieval systems as components of retrieval-augmented generation (RAG) pipelines and emphasize support for downstream question answering. Extending this line of work, Visual-RAG-ME benchmarks T2I retrieval for reasoning about visual features shared across multiple entities.

\paragraph{T2I Retrieval Methods}
A large body of work improves T2I retrieval from several angles. First, many approaches aim to train stronger multimodal embeddings through improved objectives and data mixtures \citep{DBLP:conf/bmvc/FaghriFKF18, DBLP:conf/icml/RadfordKHRGASAM21, DBLP:journals/tmlr/YuWVYSW22, DBLP:conf/icml/0001LXH22}. Other studies target specific stages of the pipeline, including textual query expansion \citep{levy2023chatting, DBLP:conf/acl/LeeYPYY24} and reranking \citep{DBLP:journals/tmlr/0002STG24, DBLP:conf/aaai/0001B00KZGL25, ding2025visual}. Finally, a recent line of work explores generative image retrieval \citep{DBLP:conf/acl/LiWQNLC24, qu2025tigerunifyingtexttoimagegeneration}, which trains a generative model to directly memorize an index of the image corpus. In contrast, VisRet expands query semantics by visualizing the query in the image space, which reduces the burden on cross-modal retrieval. It is also training-free and plug-and-play, accommodating off-the-shelf retrievers or pre-computed image embedding indices.

\section{Approach}

\subsection{Problem Formulation}

Given a textual query $q$ and an image corpus $\mathcal{I}$, \textit{Text-to-Image Retrieval} aims to retrieve $n \ge 1$ images $y_1, \ldots, y_n \in \mathcal{I}$ that best match the semantics of $q$. We further consider \textit{Visual Question Answering} (VQA), where the query is a knowledge-seeking question with an expected answer $a$.

We consider a basic retrieval-augmented generation (RAG) pipeline for VQA. A multimodal retriever $\mathcal{R}$ retrieves $k$ images from $\mathcal{I}$, denoted as $\{r_1, \ldots, r_k\} \equiv \mathcal{R}(q, \mathcal{I}) \subseteq \mathcal{I}$. A large vision-language model (LVLM) $\mathcal{M}$ then generates an answer conditioned on the question and retrieval results, i.e., $\mathcal{M}(q, \mathcal{R}(q, \mathcal{I}))$.

\subsection{Visualize-then-Retrieve}

We introduce \textit{Visualize-then-Retrieve (VisRet)}, a two-stage T2I retrieval pipeline that bridges the modality gap through modality projection. \Cref{fig:main-figure} illustrates the pipeline with an example.

\paragraph{Modality Projection}
In the first stage, VisRet uses a T2I generation system $\mathcal{T}$ to synthesize $m$ visualizations $\{v_1, \ldots, v_m\} \equiv \mathcal{T}(q)$. Concretely, given the original query, an LLM first operates in the text space to draft a T2I instruction $q'$ that describes images likely to satisfy the feature requirements implied by $q$. We then use an existing T2I generation method, such as Stable Diffusion \citep{DBLP:conf/icml/EsserKBEMSLLSBP24}, to project $q'$ into a set of image queries $\{v_1, \ldots, v_m\}$, where each $v_i$ is a synthesized image. To encourage diversity, randomness can be introduced either in the instruction drafting step or within the T2I generation process. In this work, we sample $m$ times using the same $q'$.

\paragraph{Within-Modality Retrieval}
In the second stage, VisRet performs retrieval entirely within the image modality. Each synthesized image $v_i \in \{v_1, \ldots, v_m\}$ is independently used to retrieve a ranked list of images from the corpus:
\[
\mathcal{R}(v_i, \mathcal{I}) = [r_1^{(i)}, \ldots, r_k^{(i)}].
\]
We then apply Reciprocal Rank Fusion (RRF) \citep{DBLP:conf/sigir/CormackCB09} to aggregate the $m$ ranked lists. RRF assigns each candidate image $r$ a fusion score based on its rank across the $m$ lists:
\[
\text{score}_{\text{RRF}}(r) = \sum_{i=1}^{m} \frac{1}{\lambda + \text{rank}_i(r)},
\]
where $\text{rank}_i(r)$ is the rank position of image $r$ in $\mathcal{R}(v_i, \mathcal{I})$, and $\lambda$ is a hyperparameter that controls the influence of lower-ranked items. The final top-$k$ results are obtained by selecting images with the highest $\text{score}_{\text{RRF}}(r)$.

\section{Visual-RAG-ME Benchmark}

To support the development of stronger T2I retrieval systems, we introduce \textbf{Visual-RAG-ME}, a multi-entity question answering and T2I retrieval dataset that extends Visual-RAG. Visual-RAG-ME constructs questions that compare visual features between the organism in a Visual-RAG query and another similar entity, and we manually annotate a new set of retrieval labels for the second entity. The Visual-RAG-ME annotation pipeline consists of the following three steps.

\paragraph{Second Entity Identification}
This step identifies potential entities that are biologically close to the original Visual-RAG entities, enabling plausible and challenging comparison questions. For each entity in Visual-RAG, we use BM25 \citep{10.1007/978-1-4471-2099-5_24} to retrieve up to ten entities with the most similar scientific names.

\paragraph{Query Composition and Filtering}
We (the authors) manually review all 391 questions in Visual-RAG and attempt to construct a corresponding multi-entity question. We construct a question only when we can identify images for the second entity that clearly depict the same feature as the positive images for the original entity in Visual-RAG. The questions we compose typically use a comparison format that asks whether the two organisms share a feature or which organism exhibits a more extreme stylistic feature (e.g., lighter coloration, smoother surface). This process yields 82 multi-entity questions. We then filter out (1) questions answerable by both GPT-4o and GPT-4o-mini without image information and (2) excessive questions covering the same topic. The final Visual-RAG-ME contains 50 high-quality multi-entity queries.

\paragraph{Retrieval Label Annotation and Balancing}
For each question, we collect images of the second entity from the iNaturalist 2021 dataset \citep{DBLP:conf/cvpr/HornCBWBA21} and annotate retrieval labels. We assign a positive label only when the image clearly displays the feature required to answer the question. For some questions, many positive images exist in iNaturalist, so we apply a filtering step that keeps at most 50 positive images for each entity. \Cref{tab:qualitative-examples} and Appendix \Cref{tab:additional-qualitative-examples} show example questions and their ground-truth images.

\begin{table}[t!]
\centering
\setlength{\tabcolsep}{3.2pt}
\resizebox{\linewidth}{!}{
\begin{tabular}{l|cccc}
\toprule
& \textbf{VR} & \textbf{VR-ME} & \textbf{IR-Hard} & \textbf{CC-Hard} \\
\hline
\# Queries & 391 & 50 & 59 & 500 \\
$|$Query$|$ (word count) & 18.5 & 25.1 & 6.0 & 12.2\\ 
\# Images (per entity) & 264 & 263 & 100 & 5000 \\
\# Positives (per entity) & 14.3 & 20.8 & 12.5 & 1 \\
CLIP Recall@1 & 0.210 & 0.220 & 0.000 & 0.000 \\
E5-V Recall@1 & 0.240 & 0.340 & 0.000 & 0.000 \\
\hline
\end{tabular}
}
\caption{Dataset statistics and baseline performance. VR = Visual-RAG, IR = INQUIRE-Rerank, and CC = Microsoft COCO 2017.}
\vspace{-3mm}
\label{tab:dataset-stats}
\end{table}

\Cref{tab:dataset-stats} reports the basic statistics of Visual-RAG-ME. While it contains slightly more positives per entity than Visual-RAG, it remains challenging. Both CLIP and E5-V achieve low Recall@1 due to the lengthy and knowledge-intensive queries.

\section{Main Results}
\label{section-results}

% requires \usepackage{xcolor}, \usepackage{booktabs}, \usepackage{multirow}, \usepackage{colortbl}
% \definecolor{ti}{RGB}{33,115,199}   % text->image
\definecolor{ti}{RGB}{38,111,189}   % text->image
% \definecolor{tt}{RGB}{0,150,110}    % text->text
\definecolor{tt}{RGB}{246,124,56}    % text->text
% \definecolor{ii}{RGB}{210,120,0}    % image->image
\definecolor{ii}{RGB}{63,126,49}    % image->image
\newcommand{\tirow}[1]{\textcolor{ti}{#1}}
\newcommand{\ttrow}[1]{\textcolor{tt}{#1}}
\newcommand{\iirow}[1]{\textcolor{ii}{#1}}

\setlength{\tabcolsep}{3.8pt}
\begin{table*}[ht!]
\centering
\resizebox{\linewidth}{!}{
\def\arraystretch{1.15}
\begin{tabular}{l|ccc|ccc|ccc|ccc}
\toprule
\multirow{2}{*}{\textbf{Retrieval Method}} & \multicolumn{3}{c|}{\textbf{Visual-RAG}} & \multicolumn{3}{c|}{\textbf{Visual-RAG-ME}} & \multicolumn{3}{c|}{\textbf{INQUIRE-Rerank-Hard}} & \multicolumn{3}{c}{\textbf{COCO-Hard}} \\
& \textbf{N@1} & \textbf{N@10} & \textbf{N@30} & \textbf{N@1} & \textbf{N@10} & \textbf{N@30} & \ \textbf{N@1} & \ \ \textbf{N@10} & \textbf{N@30} & \textbf{N@1} & \textbf{N@10} & \textbf{N@30} \\
\hline
\rowcolor{gray!20} \multicolumn{13}{c}{\textbf{\texttt{Retriever = CLIP}}} \\
% \hline
\tirow{\textbf{Original Query}} & 0.210 & 0.355 & 0.385 & 0.220 & 0.423 & 0.435 & \ 0.000 & \ \ 0.355 & 0.412 & 0.000 & 0.017 & 0.042 \\
\tirow{\textbf{LLM Rewriting}} & 0.238 & 0.360 & 0.395 & 0.410 & 0.575 & 0.572 & \ 0.136 & \ \ 0.349 & 0.407 & 0.014 & 0.047 & 0.093 \\
\ttrow{\textbf{Corpus Captioning (BLIP)}} & 0.105 & 0.217 & 0.271 & 0.210 & 0.354 & 0.371 & \ 0.136 & \ \ 0.319 & 0.401 & \textbf{0.054} & \textbf{0.122} & \textbf{0.153} \\
\ttrow{\textbf{VISA Reranking (top-30)}} & 0.223 & 0.378 & 0.388 & 0.240 & 0.443 & 0.457 & \ 0.000 & \ \ 0.000 & 0.000 & 0.000 & 0.000 & 0.000 \\
\ttrow{\textbf{Captioning Reranking (top-30)}} & 0.200 & 0.336 & 0.367 & 0.160 & 0.362 & 0.390 & \ 0.000 & \ \ 0.000 & 0.000 & 0.000 & 0.000 & 0.000 \\
\iirow{\textbf{\textit{VisRet}}} & \textbf{0.251} & \textbf{0.431} & \textbf{0.438} & \textbf{0.460} & \textbf{0.632} & \textbf{0.605} & \ \textbf{0.170} & \ \ \textbf{0.452} & \textbf{0.455} & 0.034 & 0.072 & 0.108 \\
\hline
\rowcolor{gray!20} \multicolumn{13}{c}{\textbf{\texttt{Retriever = E5-V}}} \\
% \hline
\tirow{\textbf{Original Query}} & 0.240 & 0.386 & 0.407 & 0.340 & 0.465 & 0.486 & \ 0.000 & \ \ 0.319 & 0.407 & 0.000 & 0.163 & 0.178 \\
\tirow{\textbf{LLM Rewriting}} & 0.223 & 0.368 & 0.391 & 0.460 & 0.569 & 0.566 & \ 0.170 & \ \ 0.367 & 0.412 & 0.038 & 0.133 & 0.182 \\
\ttrow{\textbf{Corpus Captioning (BLIP)}} & 0.156 & 0.294 & 0.319 & 0.240 & 0.405 & 0.436 & \ 0.119 & \ \ 0.342 & 0.398 & \textbf{0.064} & 0.158 & 0.196 \\
\ttrow{\textbf{VISA Reranking (top-30)}} & 0.279 & 0.413 & 0.419 & 0.350 & 0.495 & 0.500 & \ 0.022 & \ \ 0.002 & 0.002 & 0.000 & 0.000 & 0.000 \\
\ttrow{\textbf{Captioning Reranking (top-30)}} & 0.248 & 0.369 & 0.396 & 0.400 & 0.429 & 0.454 & \ 0.000 & \ \ 0.000 & 0.005 & 0.000 & 0.000 & 0.000 \\
\iirow{\textbf{\textit{VisRet}}} & \textbf{0.299} & \textbf{0.452} & \textbf{0.461} & \textbf{0.560} & \textbf{0.643} & \textbf{0.622} & \ \textbf{0.220} & \ \ \textbf{0.377} & \textbf{0.425} & 0.042 & \textbf{0.173} & \textbf{0.205} \\
\hline
\end{tabular}
}
% \vspace{-2mm}
\caption{nDCG (N@k) across four T2I retrieval benchmarks using different retrieval strategies and retrievers. Within each retriever group, the best results per column are bolded. We use different colors to denote similarity matching in \textcolor{ti}{\textbf{text-to-image}}, \textcolor{tt}{\textbf{text-to-text}}, or \textcolor{ii}{\textbf{image-to-image}} style. VisRet achieves the state-of-the-art across most evaluation settings, especially excelling in handling knowledge-intensive queries from Visual-RAG and Visual-RAG-ME. }
\label{tab:main-results-retrieval}
\vspace{-2mm}
\end{table*}

\subsection{Experimental Setup}

\paragraph{Benchmarks}
We evaluate on four benchmarks: (1) Visual-RAG \citep{visual-rag}, a T2I retrieval and VQA benchmark featuring visual knowledge-intensive questions about natural species, where the relevant evidence is often not readily available in text corpora. (2) Visual-RAG-ME, our new benchmark that compares the same visual feature across multiple entities. (3) INQUIRE-Rerank \citep{DBLP:conf/nips/VendrowPSBJABH24}, which requires accurate knowledge of species appearance and behavior for retrieval. (4) Microsoft COCO 2017 \citep{mscoco}, a widely used image captioning dataset that we repurpose for caption-to-image retrieval evaluation. For (3) and (4), we further filter queries to retain the most difficult cases, yielding INQUIRE-Rerank-Hard and COCO-Hard. \Cref{tab:dataset-stats} reports dataset statistics, and we provide additional details in Appendix \Cref{appendix-data-details}.

\paragraph{Metrics}
For all four retrieval benchmarks, we report Recall@$k$ and nDCG@$k$ with $k \in \{1, 10, 30\}$\footnote{nDCG@1 is undefined and we use Recall@1 as its value.}. For Visual-RAG and Visual-RAG-ME, we also evaluate end-to-end VQA accuracy using an LLM judge with a prompt derived from Visual-RAG \citep{visual-rag}.

\paragraph{Baselines}
We compare three families of T2I retrieval methods:
\begin{compactenum}
    \item \textbf{Text-to-Image Matching:} Retrieve using the original textual query against the image index with off-the-shelf retrievers (\textbf{CLIP}, \textbf{E5-V}). We also include a query rewriting baseline, where an LLM rewrites the query to highlight desired features before performing standard T2I retrieval over the same embedding index.
    \item \textbf{Text-to-Text Matching:} Convert images into textual surrogates and perform text-to-text retrieval avoiding cross-modal similarity matching. We compare with (1) captioning the entire corpus with BLIP, (2) top-k reranking with VISA, and (3) an ablated version of VISA, where we rerank top-k images using LLM-generated captions.
    \item \textbf{Image-to-Image Matching:} At the time of writing (December 2025), VisRet is the only method in this category.
\end{compactenum}

\paragraph{Implementation Details}
For all retrieval experiments, we use CLIP and E5-V as retrievers, \texttt{GPT-4o} \citep{OpenAI2023GPT4TR} as the T2I instruction generation LLM, and \texttt{gpt-image-1} \citep{openai_gpt_image_1} as the T2I model to generate $m=3$ images. For downstream question answering on Visual-RAG and Visual-RAG-ME, we use \texttt{GPT-4o} as the reader LLM. Full implementation details for VisRet and baselines are provided in Appendix \Cref{appendix-implementation-details}.

\subsection{VisRet Improves T2I Retrieval}
\label{section-results-retrieval}

\Cref{tab:main-results-retrieval} summarizes retrieval results across four benchmarks and two retrievers. VisRet consistently outperforms both the original-query baseline and the LLM-based query rewriting baseline. With CLIP as the retriever, VisRet improves nDCG@10 by 0.109 (38\%$\uparrow$) over the original query and by 0.064 (19\%$\uparrow$) over LLM rephrasing, averaged over four benchmarks. Similar trends hold for E5-V, where VisRet yields gains of 0.078 (23\%) and 0.052 (14\%) in nDCG@10.

Text-to-text baselines behave less consistently across benchmarks. Corpus captioning with BLIP helps on COCO, which is closest to its training distribution, but does not improve over cross-modal retrieval on benchmarks with more knowledge-intensive queries. We hypothesize that this reflects the high information density in images and the corresponding loss of potentially useful details during captioning. In contrast, reranking approaches that operate on cross-modal retrieval results, including Captioning Reranking and VISA, outperform query-agnostic corpus captioning. However, they still do not outperform direct cross-modal retrieval by a large margin. These approaches are also constrained by cost, since they process only top-$k$ candidates and their cost grows linearly with $k$. As a result, their performance is limited by the initial retrieval stage, leading to near-zero performance on INQUIRE-Rerank-Hard and COCO-Hard.

\begin{figure}[t]
\centering
\includegraphics[width=0.99\linewidth]{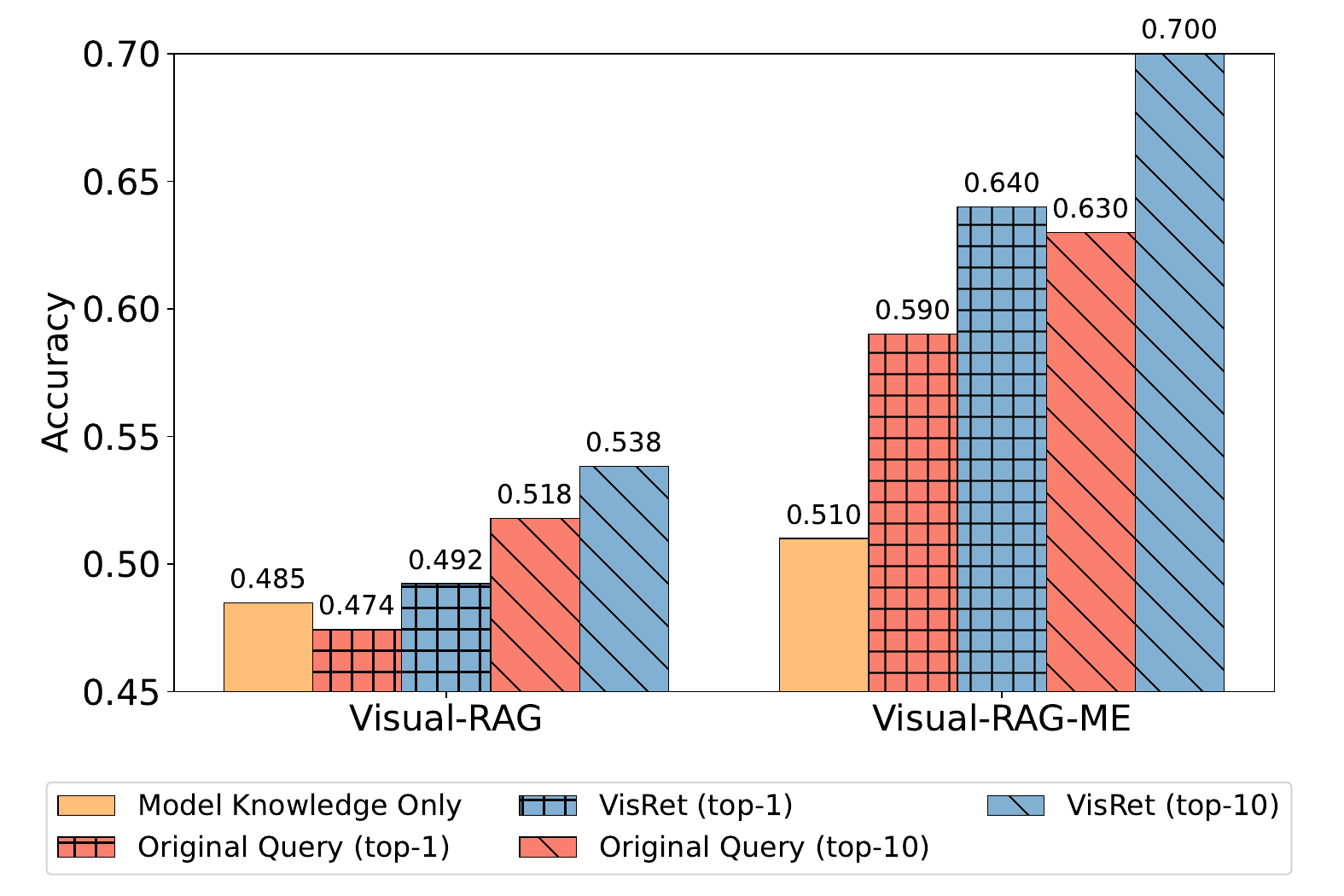}
\caption{Downstream RAG-based VQA accuracy on Visual-RAG and Visual-RAG-ME with CLIP as the retriever and GPT-4o as the reader LVLM. }
\label{fig:main-results-qa}
\end{figure}

\subsection{VisRet Improves Downstream QA}
\label{section-results-qa}

To assess VisRet in a RAG pipeline, we evaluate downstream VQA accuracy under three settings: (1) answering with only the model's internal knowledge, (2) RAG using retrieval from the original text query, and (3) RAG using VisRet. \Cref{fig:main-results-qa} reports QA accuracy on Visual-RAG and Visual-RAG-ME with GPT-4o as the LVLM reader and CLIP as the retriever. The original query often provides low-quality retrieval augmentation, and it slightly reduces performance on Visual-RAG in the top-1 retrieval setting relative to no retrieval, suggesting that inaccurate evidence can be more harmful than no evidence.

In contrast, VisRet improves QA accuracy in both top-1 and top-10 settings on both benchmarks, raising accuracy to 0.538 on Visual-RAG and 0.700 on Visual-RAG-ME. Notably, on Visual-RAG-ME, VisRet with top-1 retrieval outperforms the original-query setting even with top-10 retrieval, highlighting its ability to retrieve images that contain the required features with high precision. Overall, these results show that improvements in retrieval quality translate into tangible gains in downstream VQA performance.

\section{Analyses}

\subsection{Robustness across modules}

When VisRet is used in a RAG pipeline, four modules affect end-to-end performance: (1) visualization instruction generation, (2) query visualization, (3) retrieval, and (4) answer generation with an LVLM. We conduct analyses to probe the robustness of VisRet under variations of these modules. Overall, we find that changing the model for (1), (3), or (4) does not substantially alter VisRet's effectiveness, while performance is more sensitive to (2), which depends strongly on the capability of the T2I generation model. We first summarize the findings below:

\begin{compactenum}
   \item Using both 8B and 70B Llama \citep{grattafiori2024llama} for instruction generation instead of GPT-4o, VisRet still outperforms using the LLM itself to rephrase the query (\Cref{appendix-analsis-rephrase-model}).
   \item While most T2I models improve over cross-modality retrieval, strong proprietary T2I generation models yield substantially larger gains when used in VisRet (\Cref{tab:analysis-more-t2i-models}, \Cref{appendix-analsis-t2i-model}). We provide further discussions below.
   \item VisRet continues to deliver consistent gains when using a state-of-the-art embedding model (\Cref{tab:rzenembed-results}, \Cref{appendix-rzenembed}).
   \item Repeating the VQA experiments on Visual-RAG and Visual-RAG-ME with a stronger and a weaker reader LVLM, we find that VisRet's retrieval improvements consistently translate into downstream QA gains (\Cref{appendix-analsis-vqa-model}).
   \item Although the visualized queries improve retrieval, our preliminary study suggests they still cannot replace retrieved natural images in most cases (\Cref{appendix-analsis-gen-img-as-knowledge}).
\end{compactenum}

\begin{table}[t]
\centering
\resizebox{\linewidth}{!}{
\begin{tabular}{l|ccc|c}
\toprule
\textbf{Method} & \textbf{N@1} & \textbf{N@10} & \textbf{N@30} & \textbf{Avg} \\
\hline
\rowcolor{gray!20} \multicolumn{5}{c}{\textbf{\texttt{Retriever = CLIP}}} \\
% \hline
% \rowcolor{gray!10} \multicolumn{5}{c}{\textbf{\texttt{Baselines}}} \\
Original Query & 0.220 & 0.423 & 0.435 & 0.359 \\
\hdashline
\multicolumn{5}{l}{\textit{VisRet Variants:}}  \\
% \rowcolor{gray!10} \multicolumn{5}{c}{\textbf{\texttt{VisRet}}} \\
Stable Diffusion 3.5 & 0.270 & 0.467 & 0.484 & 0.407 \\
FLUX.1-dev & 0.320 & 0.501 & 0.494 & 0.438 \\
Qwen-Image & 0.330 & 0.501 & 0.518 & 0.450 \\
DALL-E 3 & 0.346 & 0.554 & 0.553 & 0.484 \\
Gemini & 0.410 & 0.579 & 0.583 & 0.524 \\
Image-1 (low quality) & 0.420 & 0.585 & 0.589 & 0.531 \\
Image-1 (high quality) & \textbf{0.460} & \textbf{0.632} & \textbf{0.605} & \textbf{0.566} \\
\hline
\rowcolor{gray!20} \multicolumn{5}{c}{\textbf{\texttt{Retriever = E5-V}}} \\
% \hline
% \rowcolor{gray!10} \multicolumn{5}{c}{\textbf{\texttt{Baselines}}} \\
Original Query & 0.340 & 0.465 & 0.486 & 0.430 \\
\hdashline
% \rowcolor{gray!10} \multicolumn{5}{c}{\textbf{\texttt{VisRet}}} \\
\multicolumn{5}{l}{\textit{VisRet Variants:}}  \\
Stable Diffusion 3.5 & 0.410 & 0.550 & 0.547 & 0.502 \\
FLUX.1-dev & 0.370 & 0.535 & 0.532 & 0.479 \\
Qwen-Image & 0.410 & 0.542 & 0.548 & 0.500 \\
DALL-E 3 & 0.450 & 0.599 & 0.587 & 0.545 \\
Gemini & \textbf{0.615} & 0.585 & \textbf{0.666} & 0.589 \\
Image-1 (low quality) & 0.520 & 0.629 & 0.612 & 0.587 \\
Image-1 (high quality) & 0.560 & \textbf{0.643} & 0.622 & \textbf{0.608} \\
\hline
\end{tabular}
}
\caption{nDCG@k retrieval results on Visual-RAG-ME for different T2I generation models used in VisRet. We report N@1, N@10, N@30, and their average. Best numbers within each retriever are boldfaced. Gemini stands for the Gemini-2.5-Flash-Image-Preview model. The detailed experiment setup is presented in \Cref{appendix-analsis-t2i-model}.}
\label{tab:analysis-more-t2i-models}
\end{table}

\begin{figure}[t!]
\centering
\includegraphics[width=\linewidth]{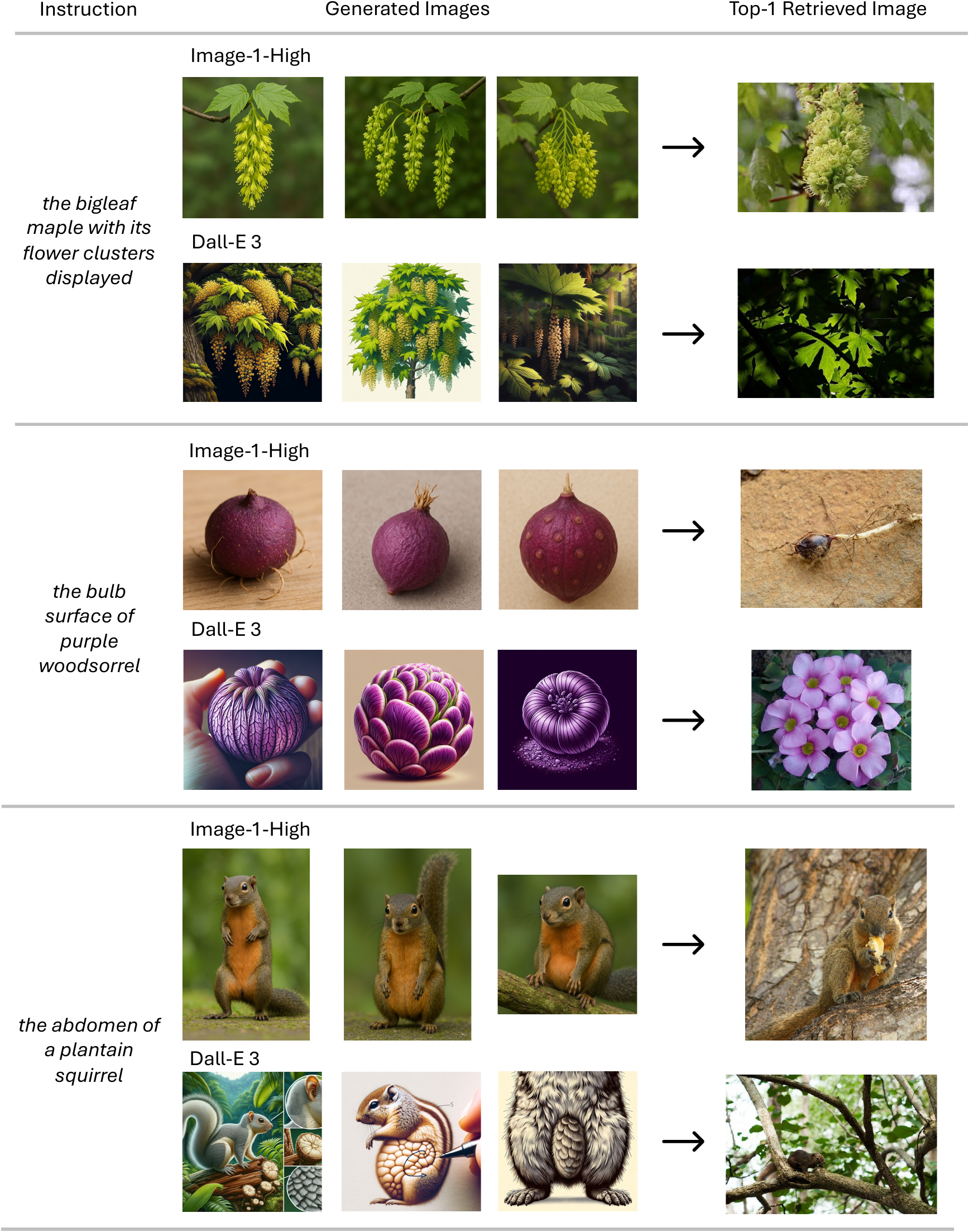}
\caption{Examples of T2I generations from Image-1 and Dall-E 3 on Visual-RAG-ME. For these examples, VisRet with Image-1 achieves a nearly perfect retrieval score while VisRet with Dall-E 3 scores close to zero.}
\label{fig:image1-vs-dalle-examples}
\end{figure}

\paragraph{T2I Model Dependency}
Despite the retrieval performance can be improved with a range of T2I generation models, VisRet is limited by the fact that it generally benefits more from stronger T2I generation models. As shown in \Cref{tab:analysis-more-t2i-models}, the latest models from Google and OpenAI dominates the performance on Visual-RAG-ME by a large margin. Setting a lower quality for OpenAI Image-1 slightly harms the performance as well. What leads to the performance gap? In \Cref{fig:image1-vs-dalle-examples}, we present three representative cases where Image-1 outperforms Dall-E 3 \citep{openai_dalle3} on Visual-RAG-ME. We find three recurring failure modes: 
\begin{itemize}
    \item \textbf{Lack of focus} As shown in the top example, Dall-E 3 fails to zoom in and highlight the key feature wanted, i.e., a flower cluster.
    \item \textbf{Factuality issues} As shown in the middle example, Dall-E 3 fails to confidently render the bulb of the organism queries.
    \item \textbf{Weak instruction following} As shown in the bottom example, Dall-E 3 fails to comply with the instruction to show the abdomen of the organism, especially in the first visualization.
\end{itemize}
Among them, factuality reflects a core capability gap, while the other two issues can often be partially mitigated through prompting. For fairness, we use the same prompt across models in \Cref{tab:analysis-more-t2i-models}, but we expect that practical deployments can further benefit from model-specific prompt tuning.

Nevertheless, we argue that this limitation does not render VisRet inapplicable in real-world application. To begin with, VisRet is in fact efficient to serve because the image embedding index can be built once and reused across queries. In contrast, reranking baselines such as VISA incur much higher per-query cost because they rely on an LVLM to process top-$k$ candidates, and their latency grows with $k$. In our experiments, VISA has roughly 5$\times$ higher latency than VisRet while reranking only the top-30 candidates. In addition, in agentic settings, VisRet can be deployed as a client-side tool that is invoked on demand for difficult queries. This avoids changes to existing retrieval serving infrastructure, reduces client-side latency, and retains VisRet's performance gains.

\subsection{Effectiveness of Repeated Runs}

\begin{figure}[t]
\centering
\includegraphics[width=\linewidth]{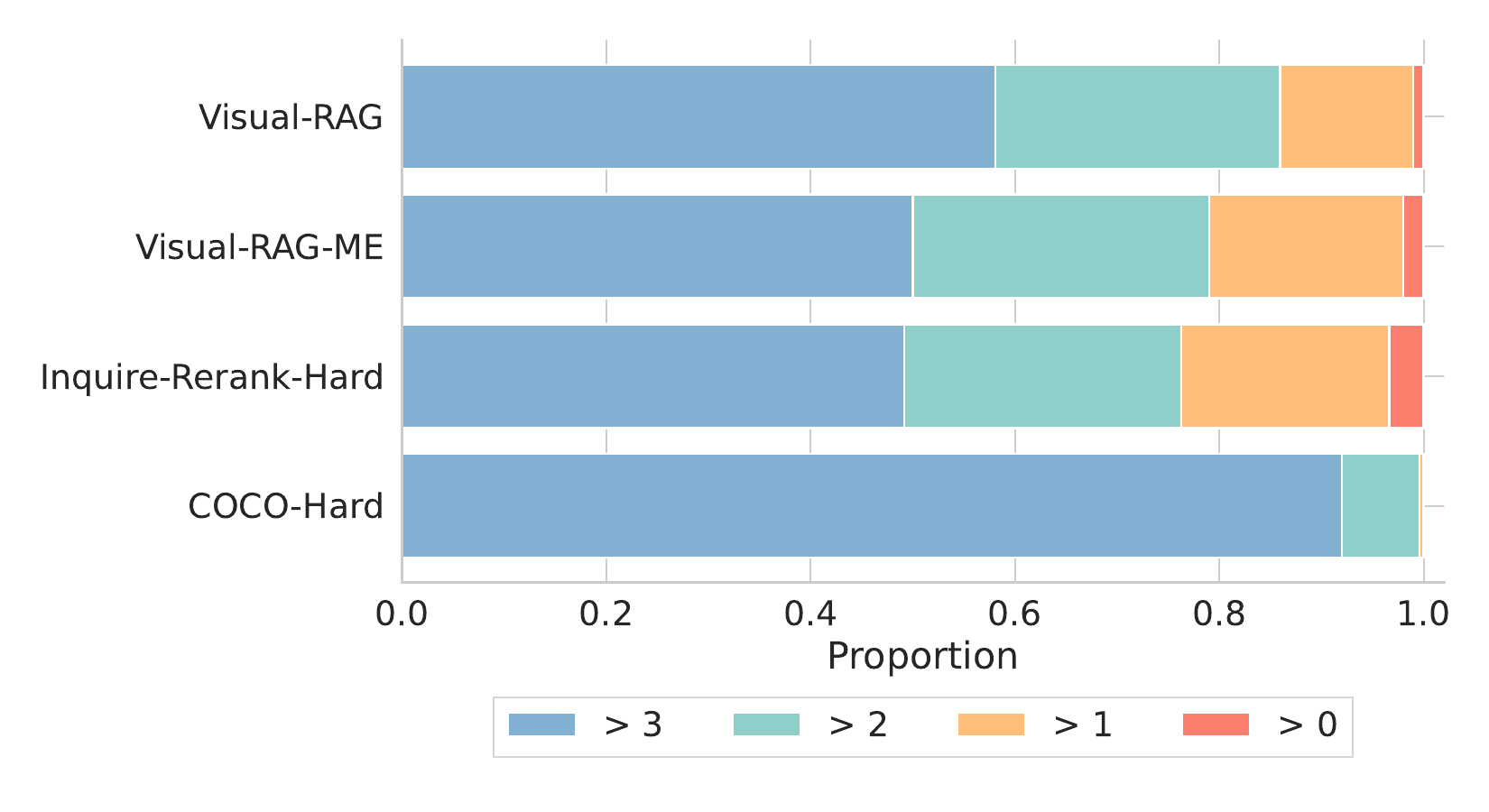}
\caption{Distribution of counts where three-image VisRet outperforming its individual samples on N@30.}
\label{fig:multi-image-vs-single-wins}
\end{figure}

Is multi-image aggregation necessary for VisRet? As shown in \Cref{fig:multi-image-vs-single-wins}, for most examples, VisRet's N@30 outperforms at least two of its individual samples, highlighting the benefit of diversifying the query through multiple visualizations. At the same time, \Cref{tab:retrieval-results-single-vs-multi} shows that using only one sample only slightly degrades overall performance. This is because in many cases, such as example 2 and 3 in \Cref{fig:image1-vs-dalle-examples}, independently generated images share substantial information overlap. In such cases, a single visualization can be an efficient option.

\begin{table}[t]
\centering
\resizebox{\linewidth}{!}{
\begin{tabular}{ll|ccc|c}
\toprule
\textbf{Benchmark} & \textbf{\# Img} & \textbf{N@1} & \textbf{N@10} & \textbf{N@30} & \textbf{Avg} \\
\hline

\rowcolor{gray!20} \multicolumn{6}{c}{\textbf{\texttt{Retriever = CLIP}}} \\
\multirow{2}{*}{Visual-RAG} & 3 & 0.251 & 0.431 & 0.438 & \textbf{0.373} \\
 & 1 & 0.243 & 0.416 & 0.425 & 0.361 \\
\hdashline
\multirow{2}{*}{Visual-RAG-ME} & 3 & 0.460 & 0.632 & 0.605 & \textbf{0.566} \\
 & 1 & 0.449 & 0.623 & 0.602 & 0.558 \\
\hdashline
\multirow{2}{*}{INQUIRE-Rerank-Hard} & 3 & 0.170 & 0.452 & 0.455 & \textbf{0.359} \\
 & 1 & 0.166 & 0.432 & 0.457 & 0.352 \\
\hdashline
\multirow{2}{*}{COCO-Hard} & 3 & 0.034 & 0.072 & 0.108 & \textbf{0.071} \\
 & 1 & 0.033 & 0.068 & 0.105 & 0.069 \\
\hline

\rowcolor{gray!20} \multicolumn{6}{c}{\textbf{\texttt{Retriever = E5-V}}} \\
\multirow{2}{*}{Visual-RAG} & 3 & 0.299 & 0.452 & 0.461 & \textbf{0.404} \\
 & 1 & 0.292 & 0.438 & 0.451 & 0.394 \\
\hdashline
\multirow{2}{*}{Visual-RAG-ME} & 3 & 0.560 & 0.643 & 0.622 & \textbf{0.608} \\
 & 1 & 0.525 & 0.638 & 0.617 & 0.593 \\
\hdashline
\multirow{2}{*}{INQUIRE-Rerank-Hard} & 3 & 0.220 & 0.377 & 0.425 & \textbf{0.341} \\
 & 1 & 0.205 & 0.292 & 0.426 & 0.308 \\
\hdashline
\multirow{2}{*}{COCO-Hard} & 3 & 0.042 & 0.173 & 0.205 & \textbf{0.140} \\
 & 1 & 0.039 & 0.167 & 0.203 & 0.136 \\
\hline
\end{tabular}
}
\caption{Retrieval results (N@k) across benchmarks and image settings for two retrievers (CLIP and E5-V). Avg is the mean of N@1, N@10, and N@30. }
\label{tab:retrieval-results-single-vs-multi}
\end{table}

\begin{table*}[t!]
\centering
\small
\resizebox{\linewidth}{!}{
\begin{tabular}{p{0.14\textwidth}p{0.22\textwidth}p{0.175\textwidth}p{0.14\textwidth}p{0.15\textwidth}p{0.13\textwidth}}
\toprule
% \midrule
\textbf{Dataset} & \textbf{Query} & \textbf{Ground Truth} & \textbf{Baseline Scores} & \textbf{Visualized Query} & \textbf{VisRet Scores} \\
\midrule
\multirow{10}*{\textbf{Visual-RAG}} & Does the Mountain Tree Frog (scientific name: Hyla eximia) have any distinctive pattern on the underside of its body?   &
\multirow{3}*{\includegraphics[width=2.7cm]{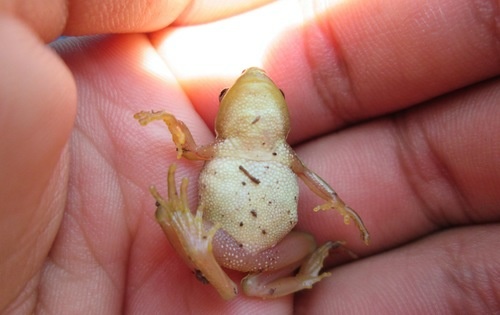}} &
\textcolor{BrickRed}{Rank:49, nDCG@10: 0.00} &
\multirow{3}*{\includegraphics[width=1.75cm]{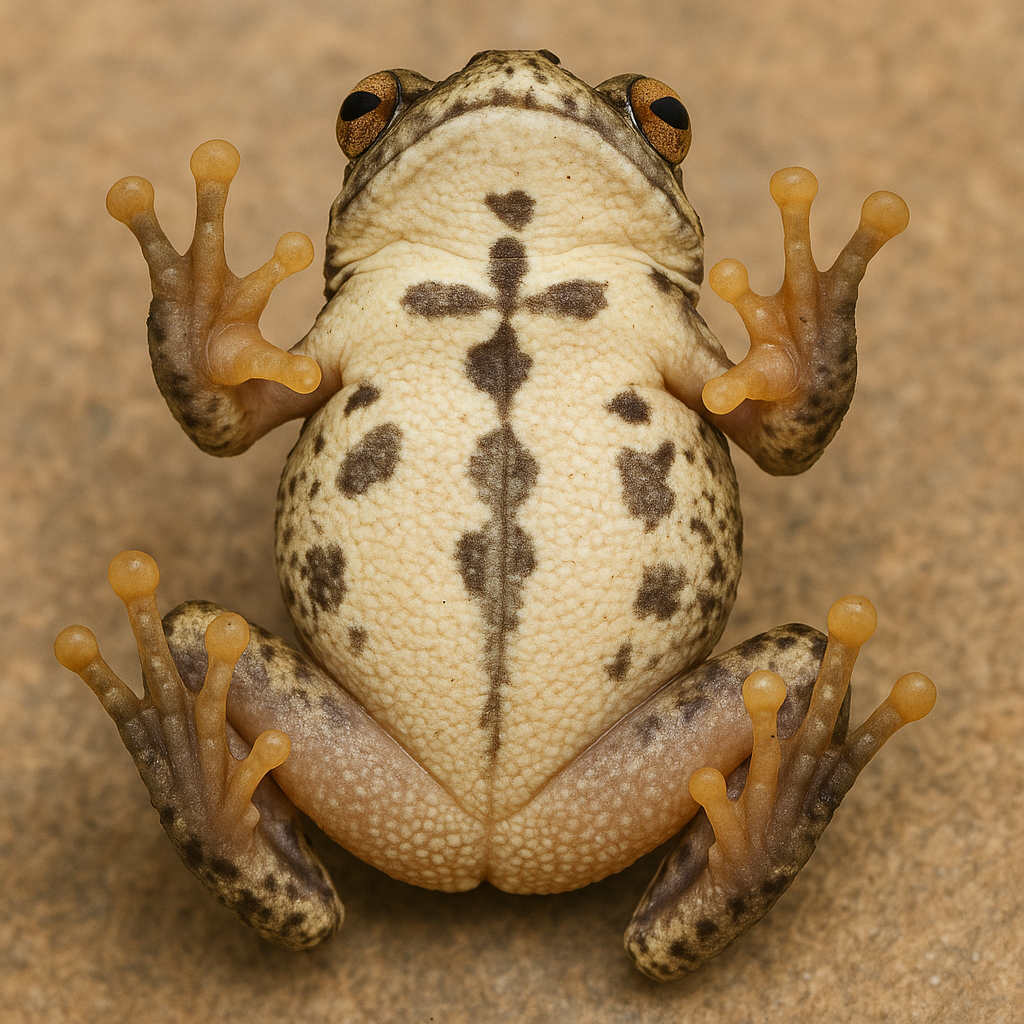}} &
\textcolor{NavyBlue}{Rank:4, nDCG@10: 0.39} \\ 
\\
\cmidrule{2-6}
 & How many petals are on each of the Tower Mustard (scientific name: Turritis glabra)'s flowers?   &
\multirow{3}*{\includegraphics[width=1.75cm]{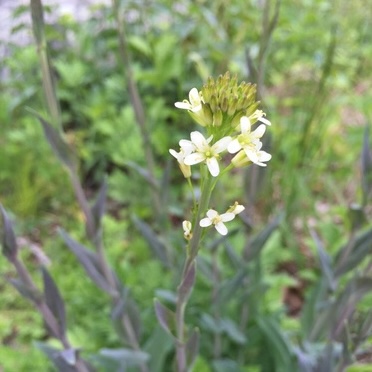}} &
\textcolor{BrickRed}{Rank:143, nDCG@10: 0.00} &
\multirow{3}*{\includegraphics[width=1.75cm]{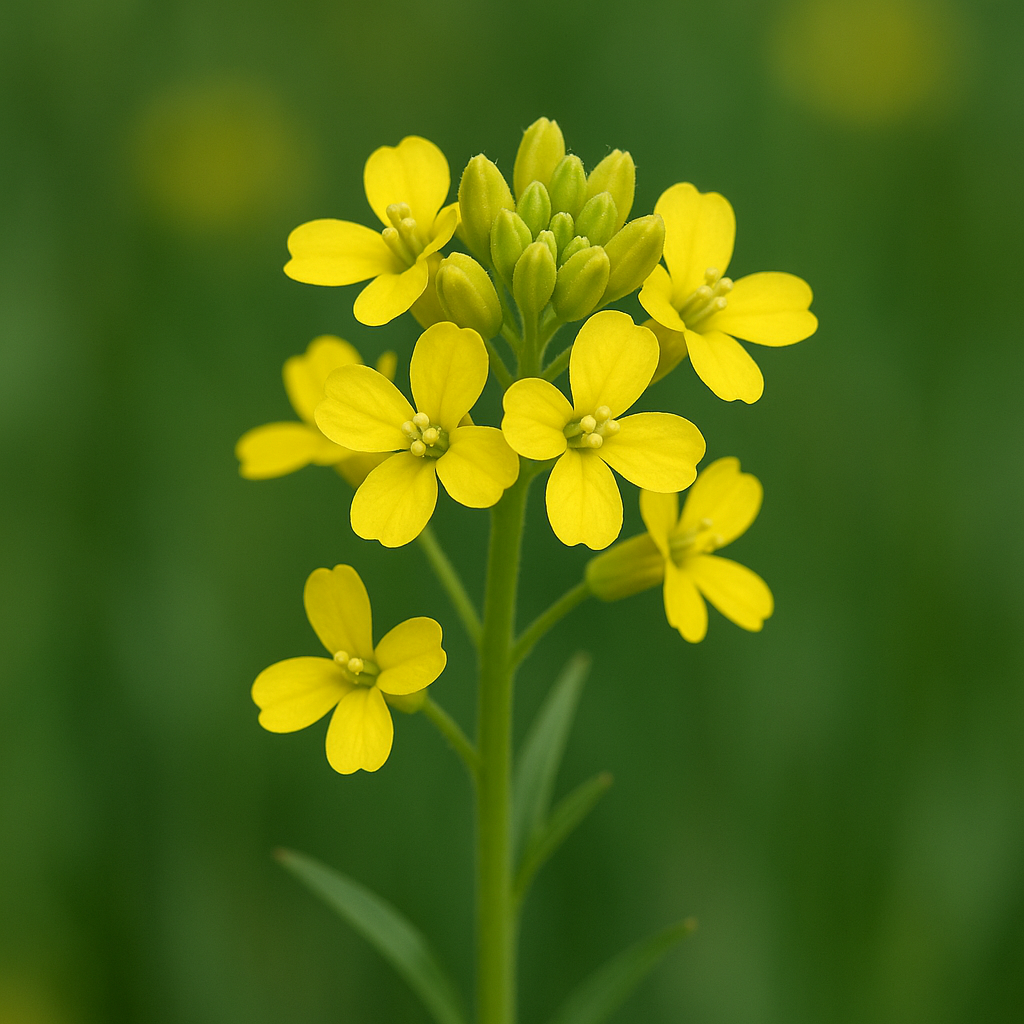}} &
\textcolor{NavyBlue}{Rank:2, nDCG@10: 0.76} \\ 
\\
\\
% \midrule
\midrule
\multirow{12}*{\textbf{Visual-RAG-ME}}
& \multirow{12}*{\parbox{1.0\linewidth}{Which one has striped primary flight feathers, Willet (scientific name: \textit{Tringa semipalmata}) or Grey-tailed tattler (scientific name: \textit{Tringa brevipes})?}}
& \multirow{3}*{\includegraphics[width=2.3cm]{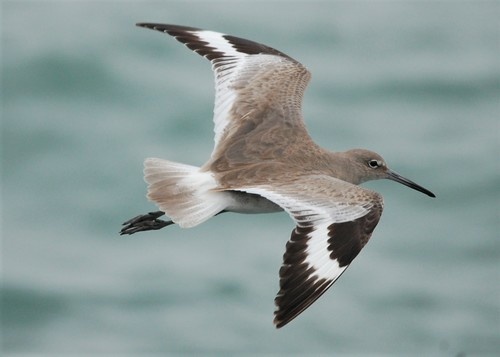}}
& Entity: Willet \textcolor{BrickRed}{Rank:104, nDCG@10: 0.00}
& \multirow{3}*{\includegraphics[width=1.75cm]{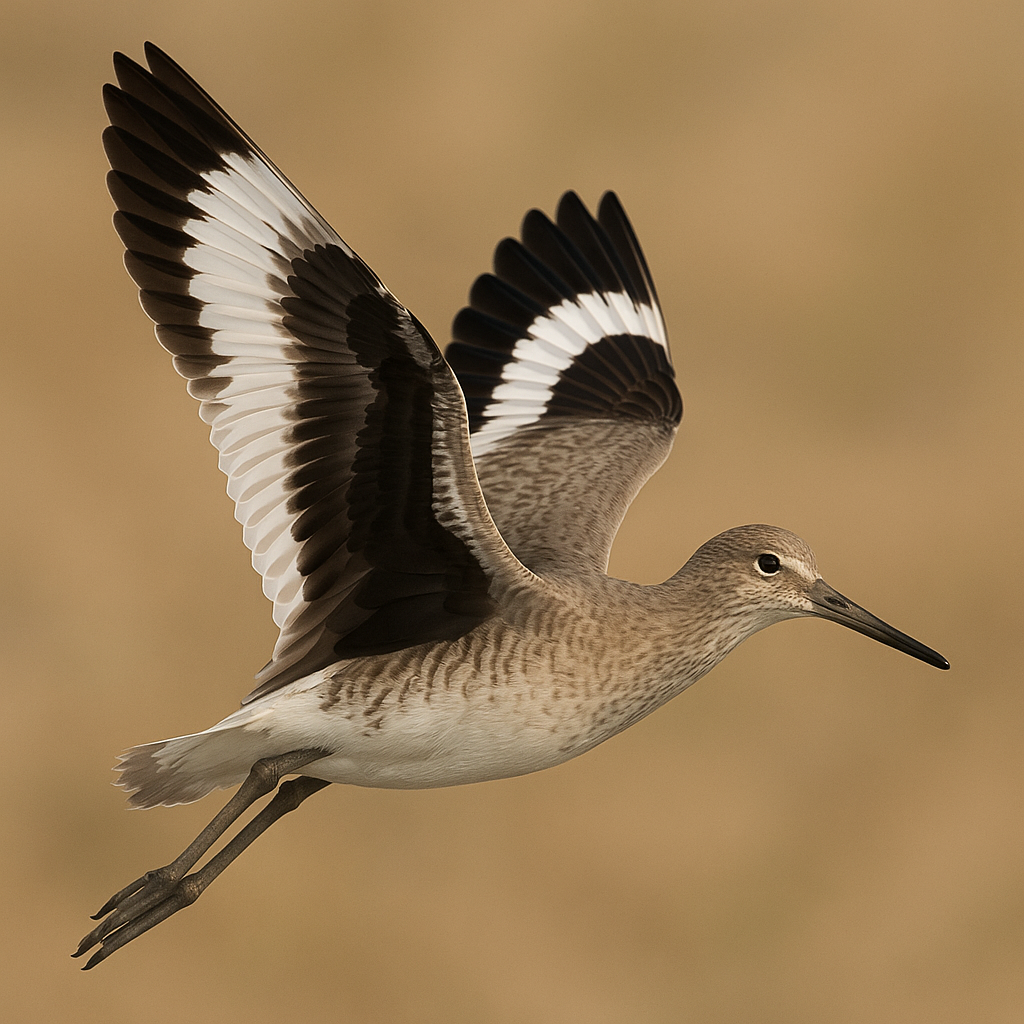}}
& Entity: Willet \textcolor{NavyBlue}{Rank:1, nDCG@10: 0.72} \\
\\
% \\
\cmidrule{3-6}
& &
\multirow{3}*{\includegraphics[width=2.3cm]{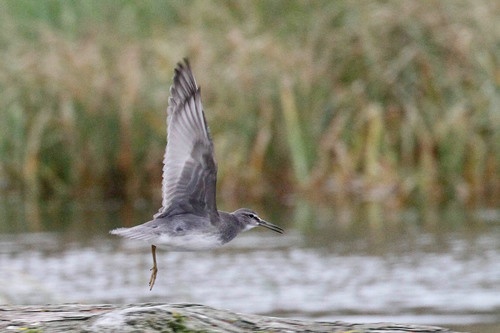}}
& Entity: Grey-tailed tattler \textcolor{BrickRed}{Rank:74, nDCG@10: 0.00}
& \multirow{3}*{\includegraphics[width=1.75cm]{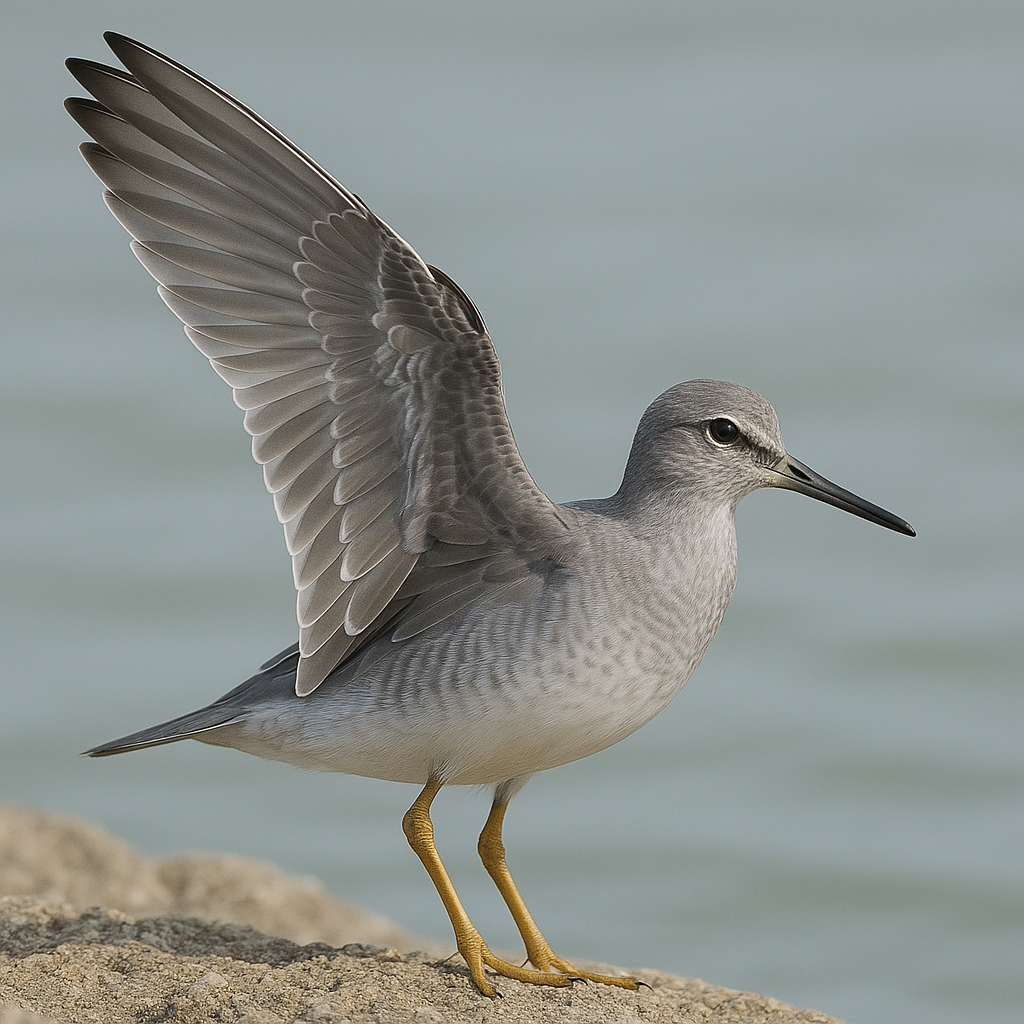}}
& Entity: Grey-tailed tattler \textcolor{NavyBlue}{Rank:1, nDCG@10: 1.00} \\
\\
\midrule
\multirow{3}*{\textbf{INQUIRE}} & A male and female cardinal sharing food  &
\multirow{3}*{\includegraphics[width=2.3cm]{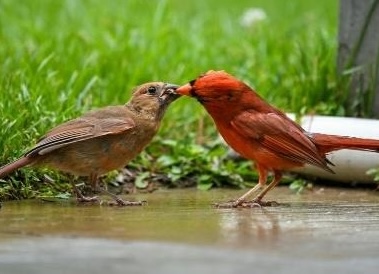}} &
\textcolor{BrickRed}{Rank:12, nDCG@10: 0.00} &
\multirow{3}*{\includegraphics[width=1.75cm]{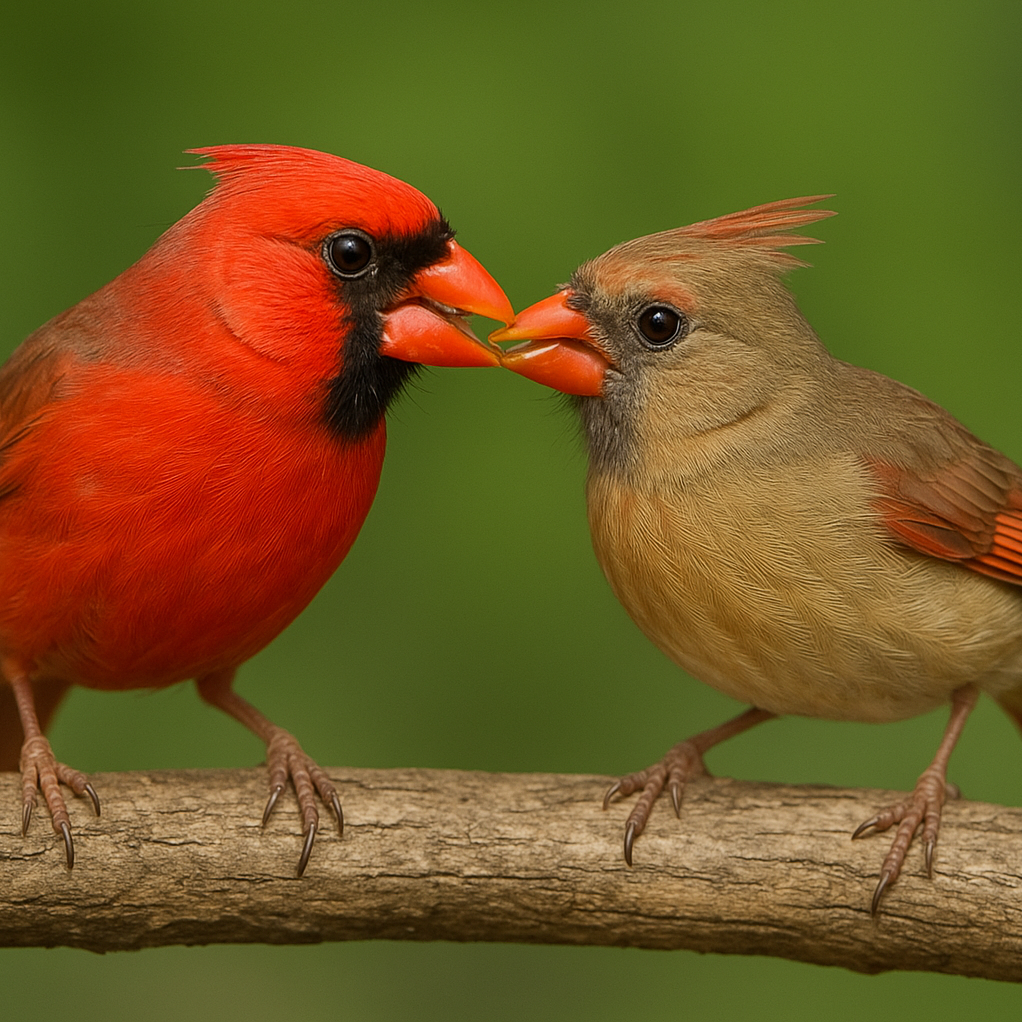}} &
\textcolor{RoyalBlue}{Rank:1, nDCG@10: 1.00} \\ 
\\
\\
\midrule
\multirow{3}*{\textbf{COCO}} & Plate with pizza, knife and fork laying on edge of plate with two glasses next to them.   & 
\multirow{3}*{\includegraphics[width=2.3cm]{figures/qual_examples/mscoco_gt24.jpg}}
& \textcolor{BrickRed}{Rank:39, nDCG@10: 0.00}
& \multirow{3}*{\includegraphics[width=1.75cm]{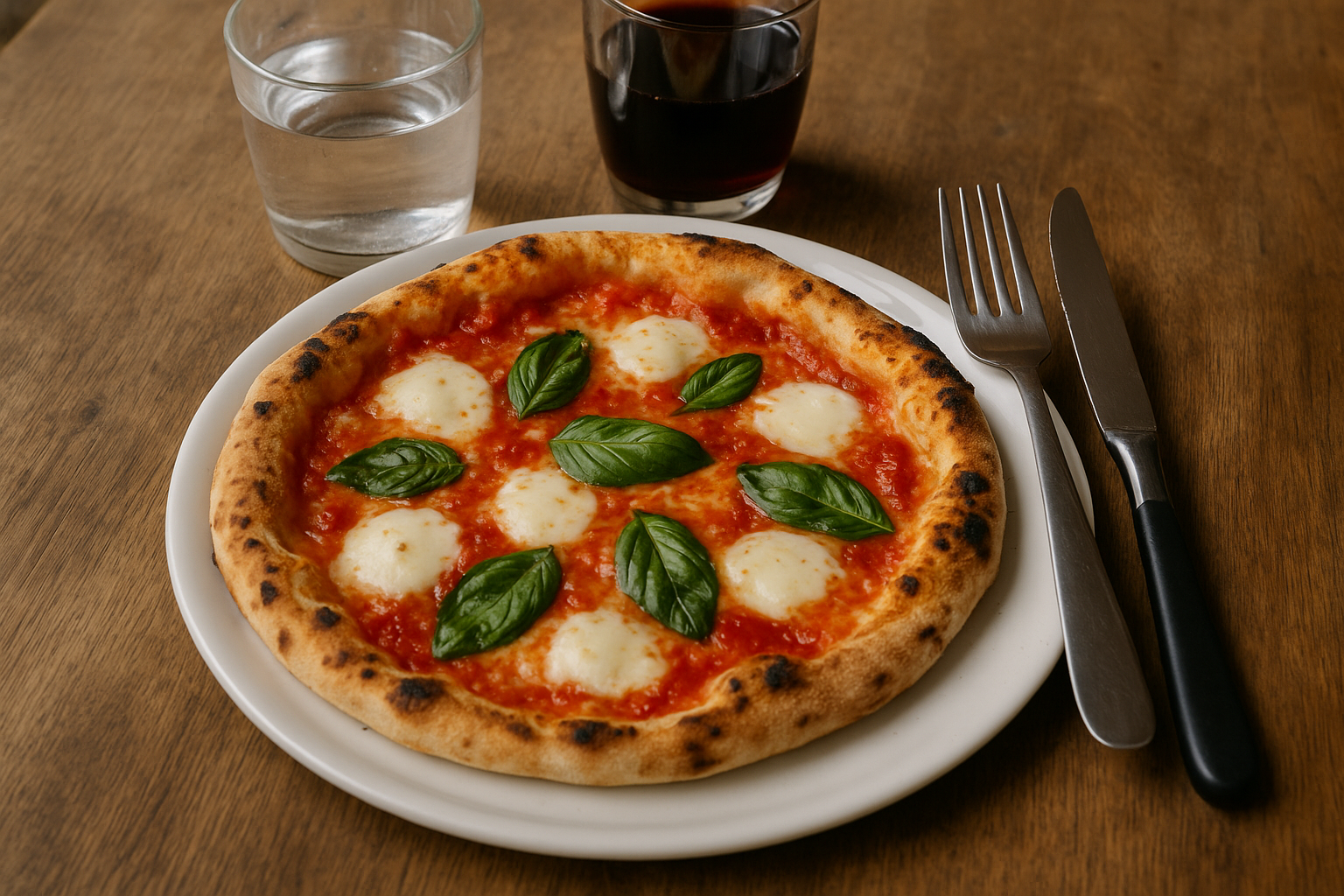}}
& \textcolor{NavyBlue}{Rank:3, nDCG@10: 0.3937} \\
% \\
\bottomrule
\end{tabular}
}
\caption{Examples: VisRet improves retrieval by highlighting visual features implied by the textual query. "Baseline" refers to cross-modal retrieval with the original query. CLIP embedding scores are used for this visualization.}
% \vspace{-4mm}
\label{tab:qualitative-examples}
\end{table*}

\begin{figure*}[ht!]
\centering
% \vspace{-2mm}
\includegraphics[width=0.99\linewidth]{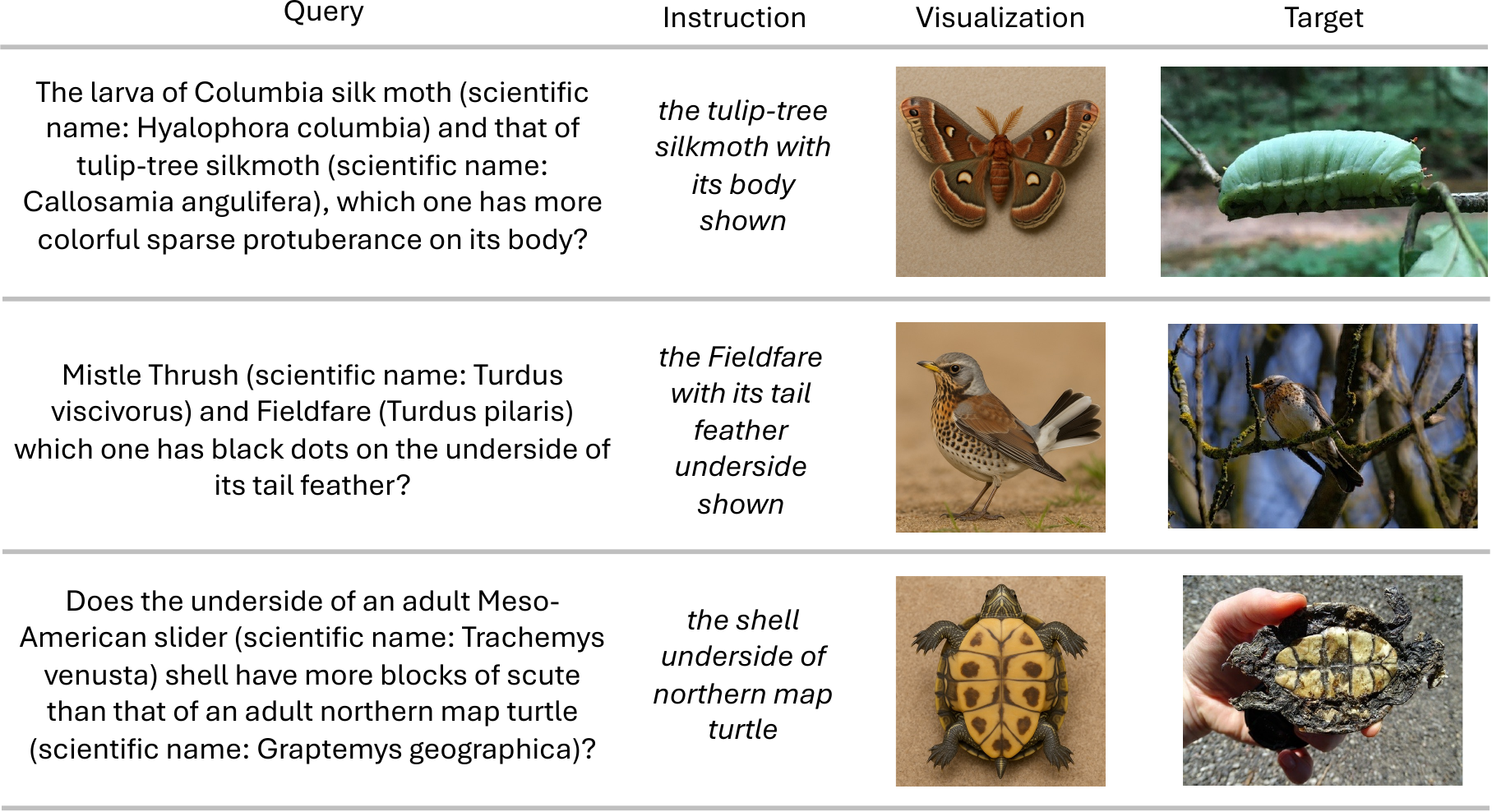}
\caption{Examples of VisRet failing to outperform the LLM query rewriting baseline. Error patterns span T2I instruction generation, visualization, and retrieval failure.}
% \vspace{-3mm}
\label{fig:visret-failure-ex}
\end{figure*}

\subsection{Success and Failure Examples}

We conclude the paper with a qualitative study. \Cref{tab:qualitative-examples} presents examples where the visualization step captures subtle visual requirements implied by the original query. Additional examples are provided in \Cref{appendix-further-qual-ex}. In many ground-truth images, the target entity appears in a specific posture that reveals the relevant evidence (Visual-RAG and Visual-RAG-ME), or it participates in complex relationships with other entities (INQUIRE and COCO). These constraints are challenging to be preserved in the embedding and surfaced during the cross-modal similarity matching. By contrast, VisRet incorporates them easily through query visualization, reducing the burden on the downstream embedding model.

In \Cref{fig:visret-failure-ex}, we show failure cases where VisRet does not outperform the LLM query rewriting baseline. We observe three salient error patterns:
\begin{itemize}
    \item \textbf{Instruction Generation}
    As shown in the top row, retrieving a natural image that contains the answer requires two features: (1) the larva form and (2) the relevant body part being shown. The generated T2I instruction omits (1) and includes only (2). Addressing this issue likely requires prompt adjustments.
    \item \textbf{Visualization}
    As shown in the second row, the instruction correctly identifies that the organism's \textit{underside} must be shown, but the T2I model fails to generate a perspective that reliably highlights the feature.
    \item \textbf{Retrieval}
    Even when instruction generation and visualization succeed, retrieval can still fail when the embedding model is sensitive to variations such as background or object appearance. In the bottom-row example, both the background (table versus human hand and ground) and the object form (alive versus dead) differ, which leads to retrieval failure.
\end{itemize}

\section{Conclusion}

This work introduces VisRet, a novel framework that visualizes text queries to reframe cross-modality T2I retrieval as within-modality retrieval. By operating entirely in the visual domain, VisRet addresses key limitations of cross-modal embedding alignment. Our experiments confirm that visualized queries substantially improve both retrieval precision and downstream VQA accuracy across four benchmark datasets and two retrievers. The simplicity and modularity of VisRet open up promising directions for future knowledge-intensive multimodal systems.

\section*{Ethics Statement}

In this section, we describe the ethical considerations related to this paper.

\paragraph{Potential Risks} Although the goal of this paper is to introduce techniques to improve the text-to-image retrieval performance, the new approach could create new social risks. Specifically, in addition to the neural embedding model, our approach involves two neural models: an LLM and a T2I generation model. It is possible for these large models to bring in new social bias in generating the visualize query and thus bias the retrieval results. For instance, when depicting certain scenes of social activity, the models could reinforce stereotypical social roles. We urge practitioners to implement model debiasing and bias detection measure when deploying our proposed T2I retrieval method in real-world applications.

\paragraph{Artifact Release} Our Visual-RAG-ME annotation is based on Visual-RAG, which is under CC BY-NC 4.0 license and the images shared by the iNaturalist 2021 dataset, which are under one of CC BY 4.0, CC BYNC 4.0, CC BY-NC-ND 4.0, CC BY-NC-SA 4.0, CC0 1.0, CC BY-ND 4.0, CC BY-SA 4.0. We adhere to the intended non-commercial research use of iNaturalist 2021 dataset and do not re-distribute the images. Following Visual-RAG, we will release our Visual-RAG-ME annotations under CC BY-NC 4.0 license.

\paragraph{Human Annotation} Two authors, who are graduate students studying Natural Language Processing, are the only annotators involved in Visual-RAG-ME annotation. Both annotators are supported by the research stipend and the annotation work counted into the working hours. Consent was obtained from both annotators before benchmark curation. The entire benchmark creation process was automatically determined exempt by the institution's IRB policy. The annotators actively discussed whenever they encounter ambiguity during annotation and reached agreements before proceeding. After the benchmark annotation, we performed a round of human auditing to ensure no question may cause privacy or ethics concerns. 

\paragraph{AI Assistant Use} AI assistants, specifically ChatGPT, are used only for revising the paper draft, fixing grammar mistakes, and improving the outlook of the figures.

\section*{Acknowledgment}

This work was partially supported by ONR grant N00014-23-1-2780, DARPA ANSR program FA8750- 23-2-0004, Taboola, Amazon, and Apple. We also thank Jia-Chen Gu and Hongwei Wang for their valuable feedback.

\bibliography{custom}

\clearpage
\appendix
\twocolumn[{%
 \centering
 \Large\bf Supplementary Material: Appendices \\ [20pt]
}]

\section{Benchmark Data Details}
\label{appendix-data-details}
In this section, we present the details of data processing for Visual-RAG, INQUIRE-Rerank-Hard, and COCO-Hard.

\subsection{Visual-RAG}

\textbf{Visual-RAG} \citep{visual-rag} releases 391 queries with associated image names from iNaturalist 2021 \citep{DBLP:conf/cvpr/HornCBWBA21} and corresponding retrieval labels\footnote{\url{https://github.com/visual-rag/visual-rag}}. To prepare the data, we download the original iNaturalist 2021 dataset and re-collect the images from the train and test sets. We were able to identify all annotated images in Visual-RAG except for a single image due to a likely path error. 

\subsection{INQUIRE-Rerank-Hard}

To prepare INQUIRE-Rerank-Hard, we accessed the publicly released INQUIRE-Rerank \citep{DBLP:conf/nips/VendrowPSBJABH24} dataset\footnote{\url{https://huggingface.co/datasets/evendrow/INQUIRE-Rerank}}. The original test set contained 160 queries, each paired with 100 images retrieved by CLIP. In a pilot study, we tested the retrieval performance of off-the-shelf CLIP and E5-V models. Results showed that CLIP can achieve 0.438 Recall@1 while E5-V achieved 0.506 Recall@1. After manually inspecting the data, we found that for many instances, the negative images are not challenging enough and it is often very straightforward to identify the ground-truth images. To highlight the challenging questions, we therefore filtered out the questions for which either CLIP or E5-V achieves a perfect Recall@1. Overall, we observe that the remaining 59 questions require more nuanced image context understanding and a higher level of knowledge of the organisms themselves, with more challenging confounding negative images. 

\subsection{COCO-Hard}

We derive COCO-Hard from the publicly released validation set of Microsoft COCO 2017 \citep{mscoco}. The original validation set contained 5000 queries, and we directly use the ground-truth images to construct the image corpus. In a pilot study, we found that CLIP is able to achieve 0.424 Recall@1 and 0.756 Recall@10, while E5-V has 0.594 Recall@1 and 0.892 Recall@10. A manual analysis shows that the queries of most examples in the original COCO dataset are short phrases centered on a single entity. To increase the task's difficulty, we rank the examples in reverse order by nDCG@30 and retain the hardest 500 instances. We find that these hard queries often describe complex spatial relationships among multiple entities.

\section{Implementation Details}
\label{appendix-implementation-details}

In this section, we present the implementation details of VisRet and baselines. 

\paragraph{VisRet: T2I Generation} To project the text query into the image space, we first instruct an LLM to analyze the query and highlight the key visual features in implies. The prompts for three benchmarking datasets are shown in \Cref{t2i-ins-prompt-visual-rag}, \Cref{t2i-ins-prompt-visual-rag-me}, and \Cref{t2i-ins-prompt-inquire}, respectively. Then, we wrap the rephrased query in a prompt template \texttt{``Generate a small image of the \{rephrased\_query\}''} to obtain the final instruction for T2I generation. We use the model \texttt{gpt-4o-2024-08-06} via OpenAI API with temperature = 0 for instruction generation and the \texttt{gpt-image-1} model for T2I generation with the quality flag set to "high". For generating multiple images for each query, we find that calling the \texttt{gpt-image-1} API to return multiple images given a single instruction already results in images with a high level of diversity. Therefore, we followed this setting in this paper and save further perturbing the instruction as future work.

\paragraph{VisRet: Retrieval} After obtaining the generated visualizations, we encode both the visualized images and the image corpus via an off-the-shelf CLIP\footnote{\url{https://huggingface.co/openai/clip-vit-large-patch14-336}} or E5-V\footnote{\url{https://huggingface.co/royokong/e5-v}} encoder and perform a similarity search. Cosine similarity is used as the similarity metric. For RRF, we use $\lambda=1$ to merge the rankings from multiple queries. All the retrieval experiments were performed on a local server with NVIDIA A100 GPU. 

\paragraph{VQA Answer Generation} We slightly modify the prompt in Visual-RAG to use chain-of-thought prompting \citep{wei2022chain}. Concretely, the model is asked to always extract visual information, perform reasoning, and conclude their reasoning with self-verification. We provide the full prompt in \Cref{qa-prompt-visrag} and \Cref{qa-prompt-visrag-me}.

\paragraph{VQA Evaluation} For Visual-RAG and Visual-RAG-ME, we use the same evaluation prompt released by the authors of Visual-RAG \citep{visual-rag}, as shown in \Cref{eval-prompt}. Since this prompt is already human-engineered for evaluating more complex references and long-form answers and the answers of Visual-RAG-ME are short and easy to evaluate, we did not perform additional prompt engineering. We use the same prompt and \texttt{gpt-4o-2024-08-06} as the LLM judge for all the VQA experiments in this paper. 

\paragraph{Baselines} For the \textbf{LLM rephrase} baseline, we use the same prompt for VisRet T2I instruction to highlight the most important features that the query is seeking. At an early stage of the project, we performed manual tuning on the prompt and found that the best-performing rephrase also serves as the best-performing T2I generation instruction. Therefore, we report the results using the same prompt for the final version for the paper. For the \textbf{Corpus Captioning} baseline, we use the \texttt{Salesforce/blip-image-captioning-large} model downloaded from huggingface. We follow the recommended parameters to run inference on local GPUs, except setting the minimum output length to 10 tokens to encourage a more detailed caption. For the \textbf{VISA Reranking} baseline, we followed the original paper's public implementation\footnote{\url{https://github.com/XLearning-SCU/2025-ICML-VISA}}, using the sample prompt except changing the model to \texttt{Qwen2.5-VL-72B-Instruct} which we empirically find performs better. For the \textbf{Captioning Reranking} baseline, we simplify the VISA pipeline by directly prompting for a caption for each of the top retrieved images using the following prompt:

\begin{center}
    \textit{"Write a caption that describes the appearance and posture of the main subject. The caption should be concise but informative so that it can be used to retrieve the images of the same subject in different postures, states, or environments." }
\end{center}

\begin{table*}[t!]
\centering
\resizebox{0.9\linewidth}{!}{
\begin{tabular}{l|c|cccccc}
\toprule
\textbf{Retrieval Strategy} & \textbf{LLM} & \textbf{R@1} & \textbf{R@10} & \textbf{R@30} & \textbf{N@1} & \textbf{N@10} & \textbf{N@30} \\
\hline
Original Query & - & 0.210 & 0.583 & 0.737 & 0.210 & 0.355 & 0.385 \\
\hdashline
\multirow{3}{*}{LLM Rewriting} 
 & Llama 3.1 8B Instruct  & 0.238 & 0.563 & 0.737 & 0.238 & 0.365 & 0.385 \\
 & Llama 3.3 70B Instruct & 0.240 & 0.575 & 0.742 & 0.240 & 0.377 & 0.399 \\
 & GPT-4o & 0.238 & 0.586 & 0.737 & 0.238 & 0.360 & 0.395 \\
 \hdashline
\multirow{3}{*}{Visualize-then-Retrieve} 
 & Llama 3.1 8B Instruct & 0.243 & 0.606 & 0.780 & 0.243 & 0.405 & 0.428 \\
 & Llama 3.3 70B Instruct & \textbf{0.256} & 0.627 & 0.790 & \textbf{0.256} & 0.413 & 0.437 \\
 & GPT-4o & 0.251 & \textbf{0.645} & \textbf{0.793} & 0.251 & \textbf{0.431} & \textbf{0.438} \\
\hline
\end{tabular}
}
\caption{Retrieval performance on Visual-RAG of with CLIP retriever, using different LLMs as T2I instruction generator for VisRet. R = Recall. N = nDCG. The best results are boldfaced.}
\label{tab:analysis-ins-gen-llms}
\end{table*}

\begin{table*}[ht]
\centering
\resizebox{\linewidth}{!}{
\begin{tabular}{l|cccc|cccc}
\toprule
\multirow{2}{*}{\textbf{Knowledge}} & \multicolumn{4}{c|}{\textbf{Visual-RAG}} & \multicolumn{4}{c}{\textbf{Visual-RAG-ME}} \\
% \cmidrule{3-5} \cmidrule{7-9}
& \# \textbf{images} & \textbf{GPT-4o-mini} &\textbf{ GPT-4o} & \textbf{GPT-4.1} & \# \textbf{images} & \textbf{GPT-4o-mini} & \textbf{GPT-4o} & \textbf{GPT-4.1} \\
\hline
Model Knowledge Only & 0 & 0.385 & 0.485 & 0.492 & 0 & 0.410 & 0.510 & 0.470 \\
\hdashline
\multirow{2}{*}{Direct T2I Retrieval}   & 1 & 0.409 & 0.474 & 0.515 & 2 & 0.490 & 0.590 & 0.610 \\
 & 10 & 0.460 & 0.518 & 0.571 & 10 & 0.480 & 0.630 & 0.650 \\
\hdashline
\multirow{2}{*}{Visualize-then-Retrieve} & 1 & 0.418 & 0.492 & \textbf{0.572} & 2 & 0.530 & 0.640 & 0.620 \\
 & 10 & \textbf{0.464} & \textbf{0.538} & 0.567 & 10 & \textbf{0.550} & \textbf{0.700} & \textbf{0.710} \\
\hline
\end{tabular}
}
\caption{VQA performance comparison using different LVLMs as instruction generators for VisRet and query rephrase models. CLIP is used as the retriever. Boldfaced numbers indicate the best in each column.}
\label{tab:analysis-qa-lvlm-choice}
\end{table*}

\begin{table*}[ht]
\centering
\resizebox{\linewidth}{!}{
\begin{tabular}{l|cccc|cccc}
\toprule
\multirow{2}{*}{\textbf{Knowledge}} & \multicolumn{4}{c|}{\textbf{Visual-RAG}} & \multicolumn{4}{c}{\textbf{Visual-RAG-ME}} \\
% \cmidrule{3-5} \cmidrule{7-9}
& \# \textbf{images} & \textbf{GPT-4o-mini} &\textbf{ GPT-4o} & \textbf{GPT-4.1} & \# \textbf{images} & \textbf{GPT-4o-mini} & \textbf{GPT-4o} & \textbf{GPT-4.1} \\
\hline
Model Knowledge Only & 0  & 0.385 & 0.485 & 0.492 & 0  & 0.410 & 0.510 & 0.470 \\
\hdashline
Generated Image (Image-1) & 1  & 0.431 & 0.425 & 0.444 & 2  & \textbf{0.590} & 0.580 & \textbf{0.800} \\
\hdashline
\multirow{2}{*}{Visualize-then-Retrieve} & 1  & 0.418 & 0.492 & \textbf{0.572} & 2  & 0.530 & 0.640 & 0.620 \\
 & 10 & \textbf{0.464} & \textbf{0.538} & 0.567 & 10 & 0.550 & \textbf{0.700} & 0.710 \\
\hline
\end{tabular}
}
\caption{VQA performance comparison using different knowledge contexts on Visual-RAG and Visual-RAG-ME. CLIP is used as the retriever. Boldfaced numbers indicate the best in each column.}
% \vspace{-2mm}
\label{tab:analysis-gen-img-as-knowledge}
\end{table*}

\section{Further Analyses}

In this section, we provide more comprehensive analyses to investigate the effectiveness of VisRet from additional perspectives, including the choices of T2I instruction LLM, T2I generation model, and the downstream VQA reader LVLM. Finally, inspired by the generative retrieval literature, we conduct a pilot study of whether the generated images could be directly used as the knowledge context for downstream question answering.

\subsection{Query Generation Model Choice}
\label{appendix-analsis-rephrase-model}

Does VisRet work well with other LLMs as the T2I instruction generator? In \Cref{tab:analysis-ins-gen-llms}, we study two differently sized open-weight LLMs for rephrasing the query and generating the T2I instruction: Llama 3.1 8B Instruct and Llama 3.3 70B Instruct \citep{grattafiori2024llama}. Overall, we observe promising results: For all three LLMs, using the LLMs themselves to generate T2I instructions for VisRet outperforms using the LLMs themselves to rephrase the query. While more expensive LLMs achieve higher performance, the small 8B Llama model can already achieve decent performance at a similar level as GPT-4o.

\setlength{\tabcolsep}{3.8pt}
\begin{table*}[ht!]
\centering
\resizebox{0.7\linewidth}{!}{
\def\arraystretch{1.15}
\begin{tabular}{l|ccc|ccc}
\toprule
\multirow{2}{*}{\textbf{Retrieval Method}} & \multicolumn{3}{c|}{\textbf{Visual-RAG}} & \multicolumn{3}{c}{\textbf{Visual-RAG-ME}} \\
& \textbf{N@1} & \textbf{N@10} & \textbf{N@30} & \textbf{N@1} & \textbf{N@10} & \textbf{N@30} \\
\hline
% \rowcolor{gray!20} \multicolumn{7}{c}{\textbf{\texttt{Retriever = RzenEmbed-v2-7B}}} \\
\tirow{\textbf{Original Query}} & 0.279 & 0.416 & 0.420 & 0.330 & 0.546 & 0.543 \\
\tirow{\textbf{LLM Rewriting}} & 0.302 & 0.414 & 0.420 & 0.390 & 0.529 & 0.560 \\
\ttrow{\textbf{Corpus Captioning (BLIP)}} & 0.128 & 0.260 & 0.310 & 0.220 & 0.375 & 0.418 \\
\ttrow{\textbf{VISA Reranking (top-30)}} & 0.288 & 0.431 & 0.442 & 0.380 & 0.547 & 0.541 \\
\ttrow{\textbf{Captioning Reranking (top-30)}} & 0.280 & 0.465 & \textbf{0.496} & 0.140 & 0.250 & 0.257 \\
\iirow{\textbf{\textit{VisRet}}} & \textbf{0.335} & \textbf{0.476} & 0.480 & \textbf{0.550} & \textbf{0.667} & \textbf{0.642} \\
\hline
\end{tabular}
}
\caption{Additional evaluation results for \texttt{Retriever = RzenEmbed-v2-7B}. nDCG (N@k) is reported. VisRet still strongly outperforms most baselines in most settings except captioning reranking for Visual-RAG's N@30.}
\label{tab:rzenembed-results}
% \vspace{4mm}
\end{table*}

\subsection{T2I Generation Model Choice}
\label{appendix-analsis-t2i-model}

How strong does the T2I generation model need to be for VisRet to work well? We compare the default T2I generation model (Image-1 with high quality setting) with several other commonly used, strong T2I generation models:
\begin{itemize}
    
    \item Stable Diffusion 3 \citep{DBLP:conf/icml/EsserKBEMSLLSBP24}.
    \item FLUX-1.dev \citep{bfl-announcing-2024}.
    \item Qwen-Image \citep{wu2025qwenimagetechnicalreport}.
    \item DALL-E 3 API \citep{openai_dalle3}.
    \item  Gemini-2.5-Flash-Image-Preview API \citep{fortin2025-gemini25-flash-image}.
    \item Image-1 API  with the low quality setting.
\end{itemize}

\Cref{tab:analysis-more-t2i-models} shows the results with GPT-4o as the T2I instruction generation model and CLIP as the retriever model. We observe that while all the models can improve over cross-modal retrieval to some extent in the VisRet pipeline, their effectiveness increases with image quality. The best performance is achieved by the latest proprietary models Image-1 and Gemini-2.5-Flash-Image-Preview. Together, these results suggest that a good T2I generation model with strong instruction-following ability is necessary for VisRet. As further T2I generation methods improve, we anticipate that building more cost-efficient versions of VisRet is viable and promising.

\subsection{Downstream VQA Model Choice}
\label{appendix-analsis-vqa-model}

While we have shown the benefit of VisRet for VQA for GPT-4o, does the improvement hold across LVLMs with different capabilities? In \Cref{tab:analysis-qa-lvlm-choice}, we repeat the VQA experiments with two additional LVLMs: GPT-4o-mini (version \texttt{gpt-4o-mini-2024-07-18}) and GPT-4.1 (version \texttt{gpt-4.1-2025-04-14}). Overall, we observe similar trends as those presented in \Cref{fig:main-results-qa}. Both direct T2I retrieval and VisRet outperform relying only on the model's knowledge, with VisRet substantially outperforming the former. These results form the foundation for VisRet as a general plug-and-play method to enhance RAG pipelines that rely on accurate T2I retrieval.

\subsection{Image Queries as Knowledge}
\label{appendix-analsis-gen-img-as-knowledge}

As demonstrated by previous results, a T2I generation model with strong ability to follow instructions and generate realistic images is crucial to the success of VisRet. A natural question is whether it is still necessary to perform retrieval rather than directly using the generated images as the knowledge? In \Cref{tab:analysis-gen-img-as-knowledge}, we compare the performance of using a single image as the context with VisRet. Overall, we observe mixed results. For Visual-RAG, GPT-4o-mini achieves slightly higher performance with the generated image than top-1 retrieval, but GPT-4o and GPT-4.1 exhibit the reverse pattern. For Visual-RAG-ME, both GPT-4o-mini and GPT-4.1 prefer the generated image to top-1 retrieval (and even top-10 retrieval). However, when provided with top-10 retrieved images, the models generally exhibit higher VQA performance than when using the generated image. Therefore, we conclude that retrieving natural images is still crucial for challenging VQA tasks like Visual-RAG and Visual-RAG-ME and cannot be fully replaced by pure T2I generation at this stage. An important direction for future work is to combine image generation and image retrieval to improve the quality of the retrieved knowledge.

\subsection{Stronger Embedding Models}
\label{appendix-rzenembed}

In \Cref{tab:rzenembed-results}, we provide additional results using RzenEmbed-v2-7B\footnote{\url{https://huggingface.co/qihoo360/RzenEmbed}}, which is one of the best models on MMEB at the time of writing (2025 December). VisRet still strongly outperforms most baselines in most settings except captioning reranking for Visual-RAG’s N@30. This result highlights both weaker and stronger embedding models benefit from query visualization.

\begin{table*}[t!]
\centering
\small
\vspace{-4mm}
\resizebox{\linewidth}{!}{
\begin{tabular}
{p{0.15\textwidth}p{0.28\textwidth}p{0.155\textwidth}p{0.125\textwidth}p{0.15\textwidth}p{0.125\textwidth}}
\toprule
% \midrule
\textbf{Dataset} & \textbf{Query} & \textbf{Ground Truth} & \textbf{Baseline Scores} & \textbf{Visualized Query} & \textbf{VisRet Scores} \\
\midrule
\multirow{15}*{\textbf{Visual-RAG}} & What is the color of the head of larva Urania Swallowtail Moth (scientific name: Urania fulgens)?   &
\multirow{3}*{\includegraphics[width=2.2cm]{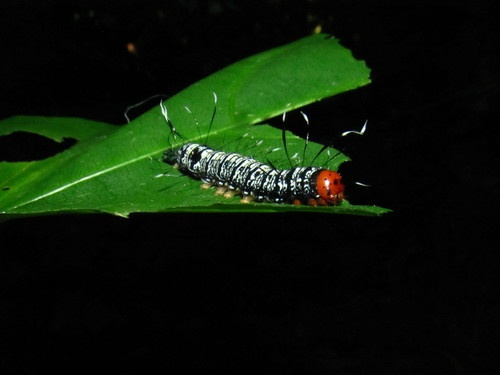}} &
\textcolor{BrickRed}{Rank:96, nDCG@10: 0.00} &
\multirow{3}*{\includegraphics[width=1.65cm]{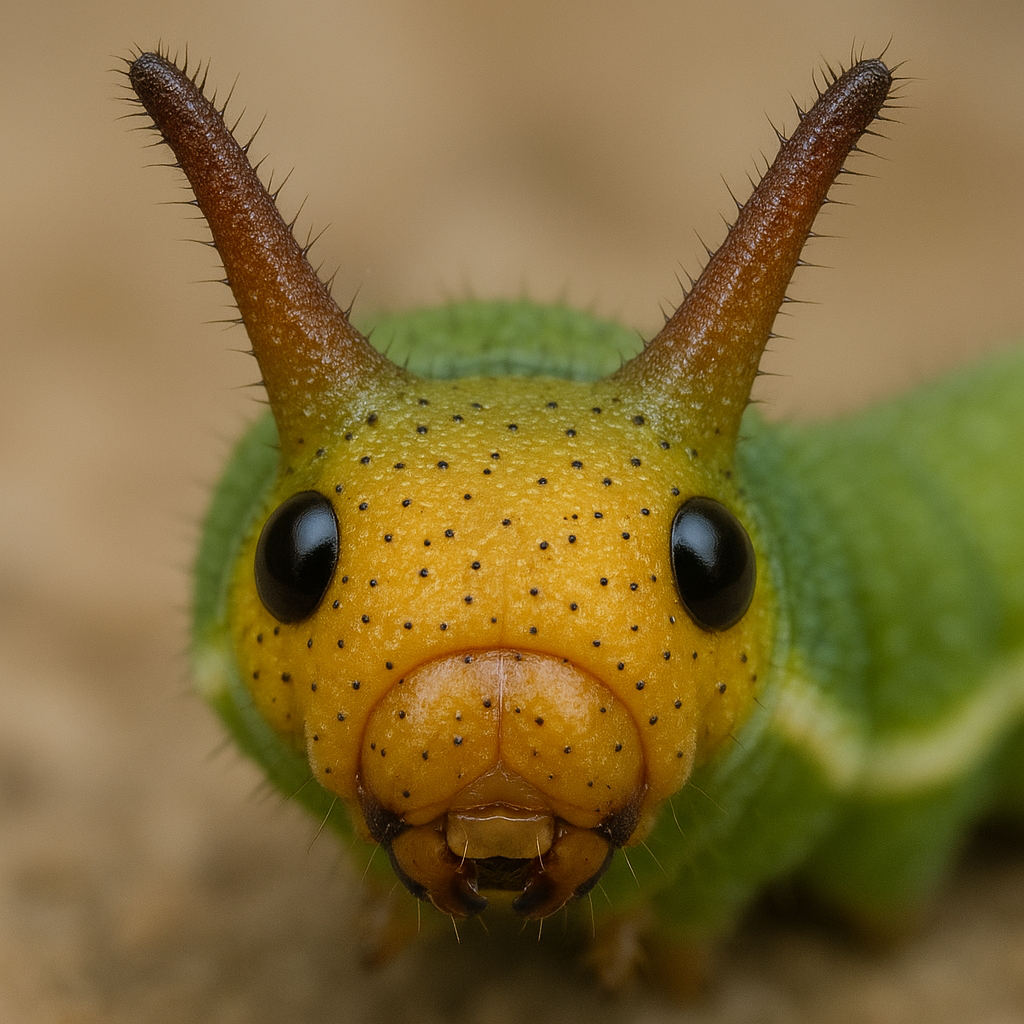}} &
\textcolor{NavyBlue}{Rank:1, nDCG@10: 0.66} \\ 
\\
\\ 
\cmidrule{2-6}
 & What color are the ventral abdomen of Golden Buprestid Beetle (scientific name: Buprestis aurulenta)?   &
\multirow{3}*{\includegraphics[width=2.25cm]{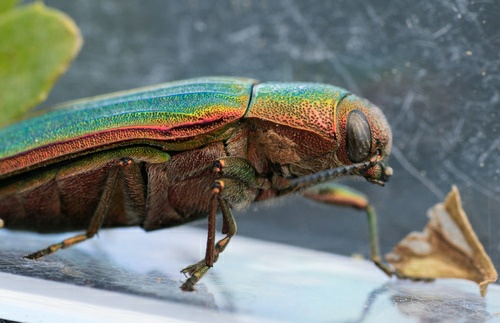}} &
\textcolor{BrickRed}{Rank:23, nDCG@10: 0.00} &
\multirow{3}*{\includegraphics[width=1.65cm]{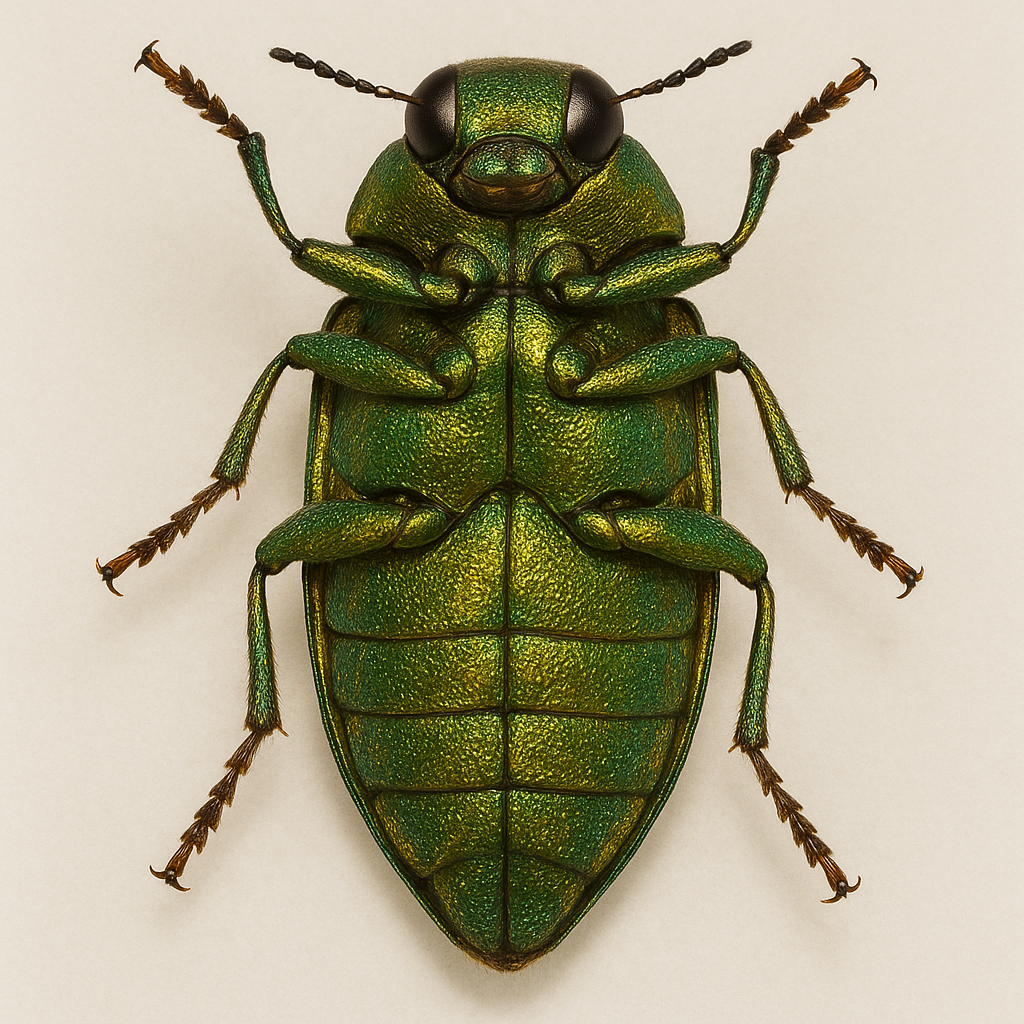}} &
\textcolor{NavyBlue}{Rank:4, nDCG@10: 0.39} \\ 
\\
\\
\cmidrule{2-6}
& Are there visually distinctive scales on feet of Azure Tit (scientific name: Cyanistes cyanus)?   &
\multirow{3}*{\includegraphics[width=1.7cm]{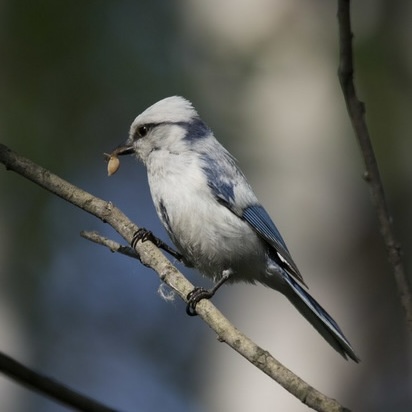}} &
\textcolor{BrickRed}{Rank:26, nDCG@10: 0.40} &
\multirow{3}*{\includegraphics[width=1.7cm]{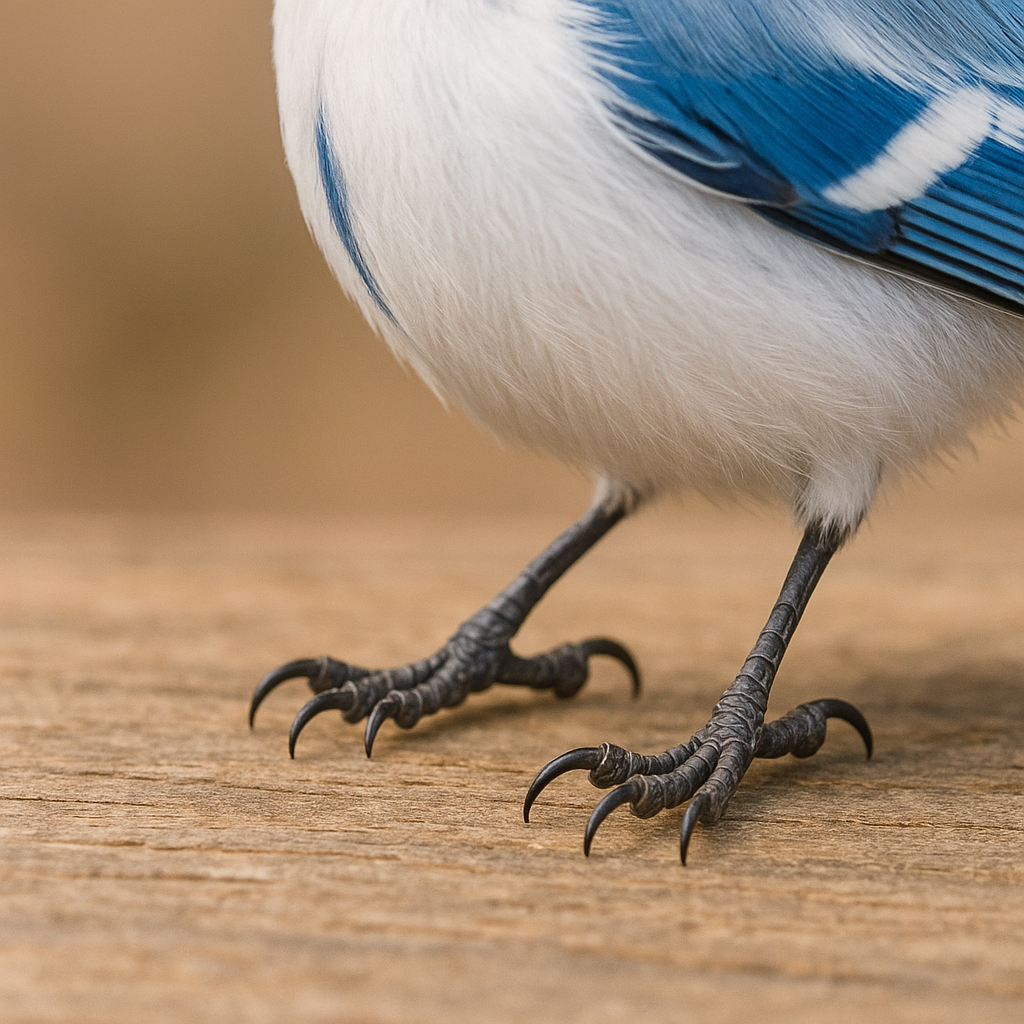}} &
\textcolor{NavyBlue}{Rank:2, nDCG@10: 0.76} \\ 
\\
\\
\midrule
% \multirow{20}*{\textbf{INQUIRE}}& A male and female cardinal sharing food  &
% \multirow{3}*{\includegraphics[width=2.3cm]{figures/qual_examples/inquire_gt1.jpg}} &
% \textcolor{BrickRed}{Rank:12, nDCG@10: 0.00} &
% \multirow{3}*{\includegraphics[width=1.65cm]{figures/qual_examples/inquire_gen1.png}} &
% \textcolor{NavyBlue}{Rank:1, nDCG@10: 1.00} \\ 
% \\
% \\
% \cmidrule{2-6}
\multirow{15}*{\textbf{INQUIRE}} & Mexican grass-carrying wasp visiting a purple flower   &
\multirow{3}*{\includegraphics[width=2.4cm]{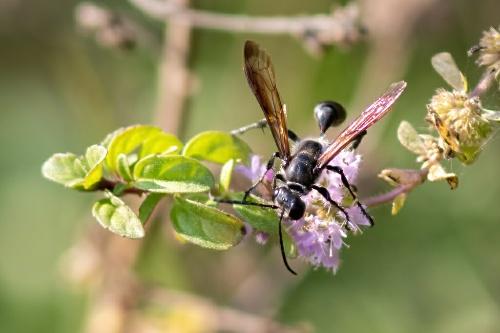}} &
\textcolor{BrickRed}{Rank:45, nDCG@10: 0.00} &
\multirow{3}*{\includegraphics[width=1.65cm]{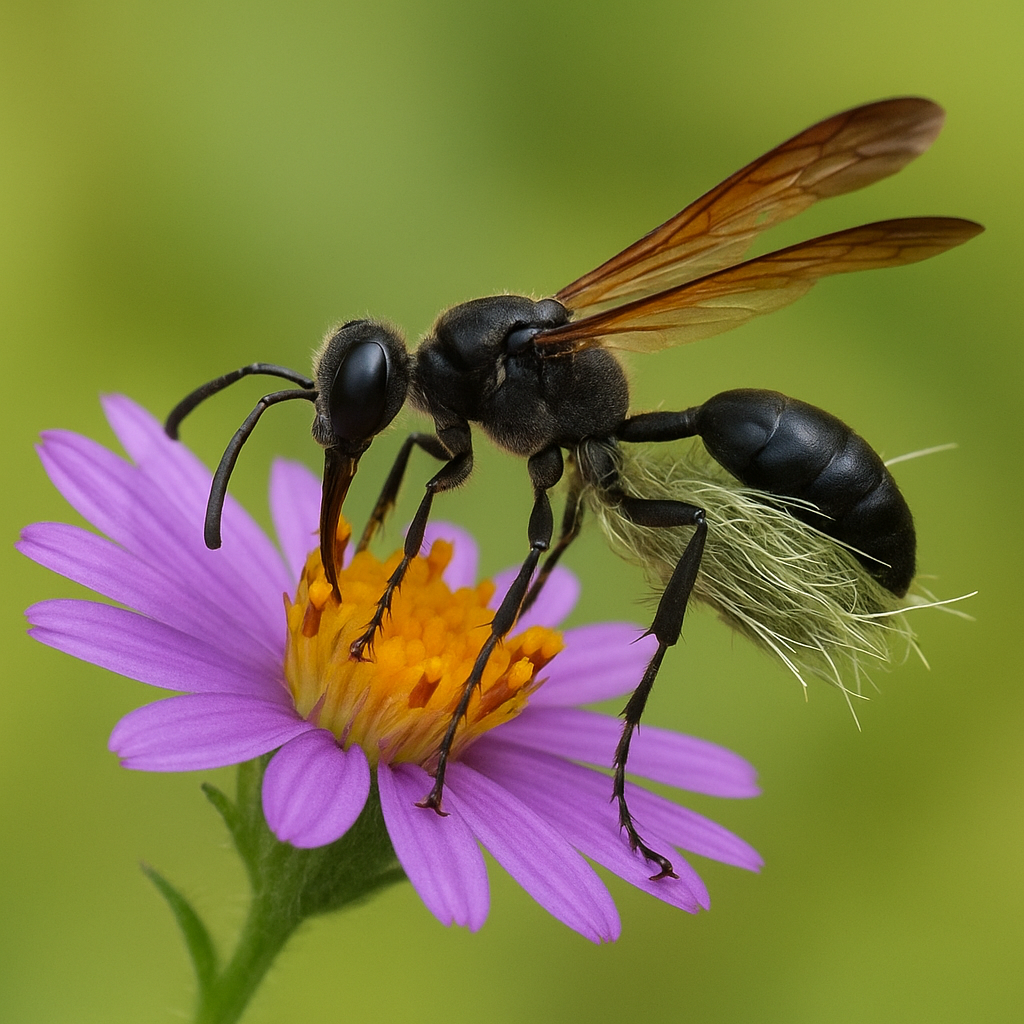}} &
\textcolor{NavyBlue}{Rank:4, nDCG@10: 0.39} \\ 
\\
\\
\cmidrule{2-6}
 & great golden digger wasp carrying an orthopteron    &
\multirow{3}*{\includegraphics[width=2.4cm]{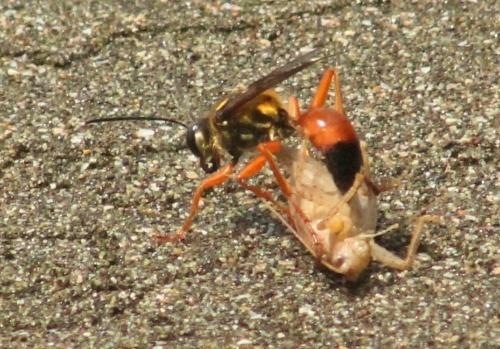}} &
\textcolor{BrickRed}{Rank:12, nDCG@10: 0.45} &
\multirow{3}*{\includegraphics[width=1.65cm]{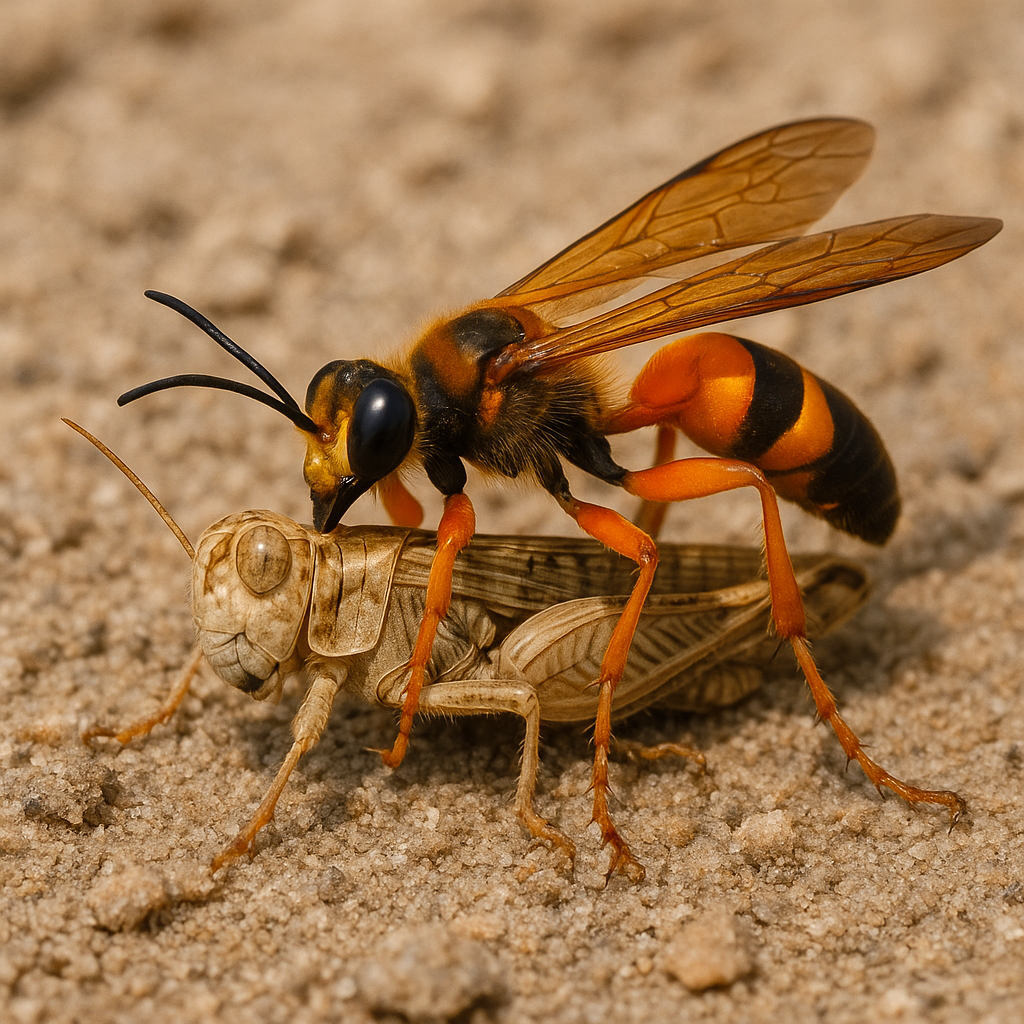}} &
\textcolor{NavyBlue}{Rank:2, nDCG@10: 0.83} \\ 
% \\
% \\
% \cmidrule{2-6}
%   & male ruby-throated hummingbird in flight     &
% \multirow{3}*{\includegraphics[width=2.4cm]{figures/qual_examples/inquire_gt4.jpg}} &
% \textcolor{BrickRed}{Rank:60, nDCG@10: 0.00} &
% \multirow{3}*{\includegraphics[width=1.65cm]{figures/qual_examples/inquire_gen4.png}} &
% \textcolor{NavyBlue}{Rank:7, nDCG@10: 0.39} \\ 
\\
\\
\midrule
\multirow{20}*{\textbf{Visual-RAG-ME}} & \multirow{10}*{\parbox{1.0\linewidth}{Which one has striped primary flight feathers, Willet (scientific name: Tringa semipalmata) or Grey-tailed tattler (scientific name: Tringa brevipes)?}}   &
\multirow{3}*{\includegraphics[width=2.4cm]{figures/qual_examples/me_gt1_e1.jpg}} &  Entity: Willet 
\textcolor{BrickRed}{Rank:104, nDCG@10: 0.00} &
\multirow{3}*{\includegraphics[width=1.65cm]{figures/qual_examples/me_gen1_e1.png}} & Entity: Willet
\textcolor{NavyBlue}{Rank:1, nDCG@10: 0.72} \\ 
\\
\cmidrule{3-6}
& & 
\multirow{3}*{\includegraphics[width=2.4cm]{figures/qual_examples/me_gt1_e2.jpg}} & Entity: Grey-tailed tattler 
\textcolor{BrickRed}{Rank:74, nDCG@10: 0.00} &
\multirow{3}*{\includegraphics[width=1.65cm]{figures/qual_examples/me_gen1_e2.png}} & Entity: Grey-tailed tattler 
\textcolor{NavyBlue}{Rank:1, nDCG@10: 1.00} \\ 
\\
\cmidrule{2-6}
& \multirow{10}*{\parbox{1.0\linewidth}{ Which one has less prominent color patterns, silvery checkerspot (scientific name: Chlosyne nycteis) caterpillar or theona checkerspot (scientific name: Chlosyne theona) caterpillar? }}   &
\multirow{3}*{\includegraphics[width=2.4cm]{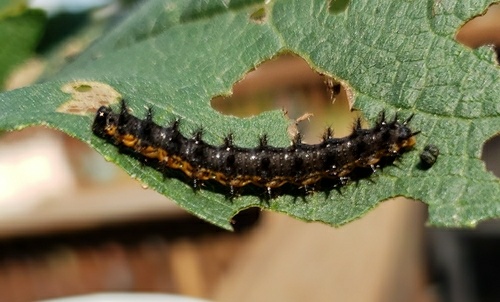}} & Entity: silvery checkerspot
\textcolor{BrickRed}{Rank:188, nDCG@10: 0.45} &
\multirow{3}*{\includegraphics[width=1.65cm]{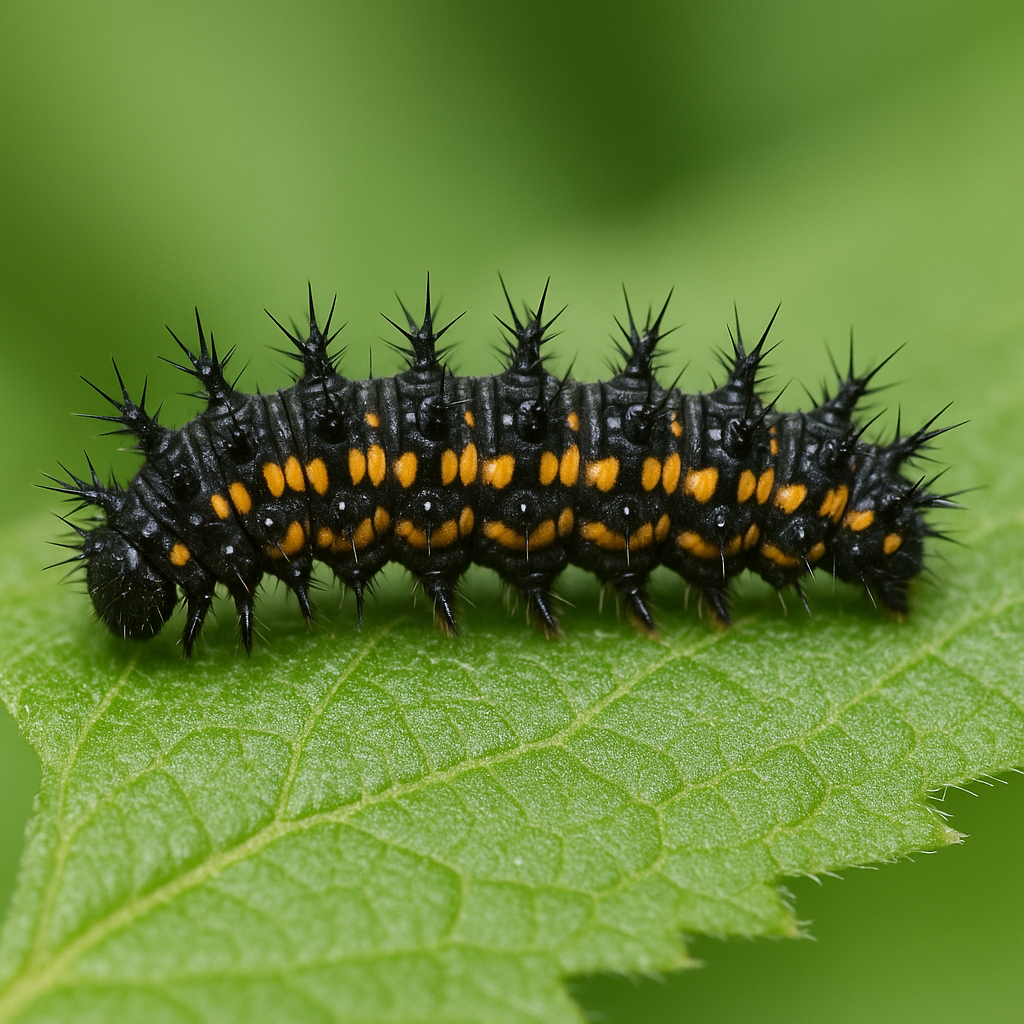}} & Entity: silvery checkerspot
\textcolor{NavyBlue}{Rank:4, nDCG@10: 0.88} \\ 
\\
\cmidrule{3-6}
& & 
\multirow{3}*{\includegraphics[width=2.4cm]{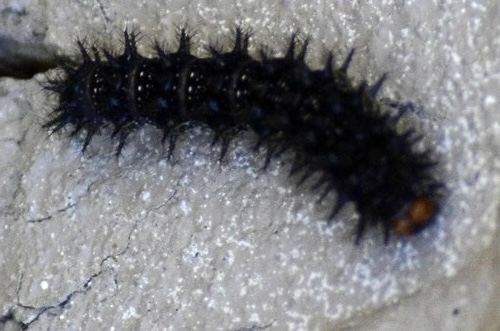}} &
Entity: theona checkerspot
\textcolor{BrickRed}{Rank:120, nDCG@10: 0.52} &
\multirow{3}*{\includegraphics[width=1.65cm]{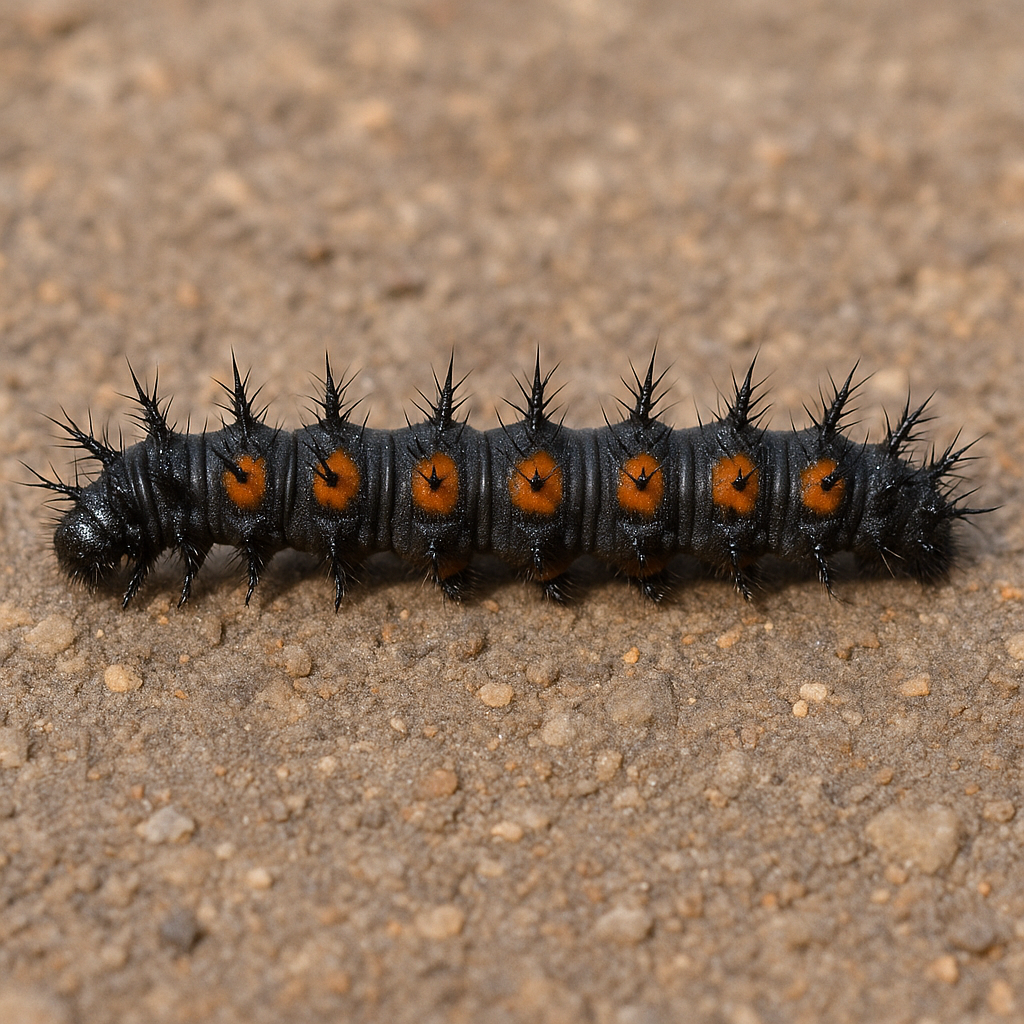}} &
Entity: theona checkerspot
\textcolor{NavyBlue}{Rank:3, nDCG@10: 0.85} \\ 
\midrule
\multirow{15}*{\textbf{COCO}} & A den with a large coffee table, couch and a television.   &
\multirow{3}*{\includegraphics[width=2.4cm]{figures/qual_examples/mscoco_gt6.jpg}} &
\textcolor{BrickRed}{Rank:61, nDCG@10: 0.00} &
\multirow{3}*{\includegraphics[width=1.65cm]{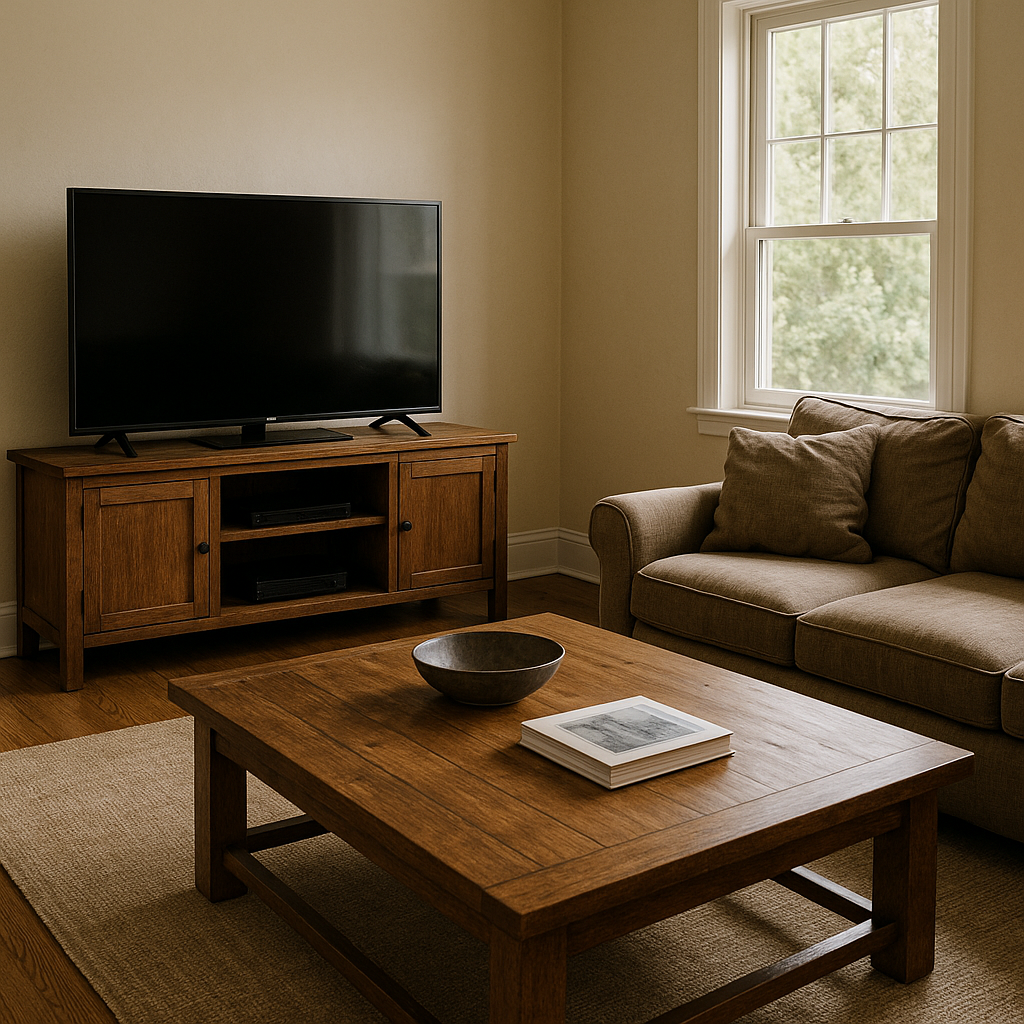}} &
\textcolor{RoyalBlue}{Rank:4, nDCG@10: 0.3937} \\ % this is visret with image-1-high, 3 imgs
\\
\\
\cmidrule{2-6}
& A computer workstation with a laptop and a desktop computer.   &
\multirow{3}*{\includegraphics[width=2.4cm]{figures/qual_examples/mscoco_gt64.jpg}} &
\textcolor{BrickRed}{Rank:37, nDCG@10: 0.00} &
\multirow{3}*{\includegraphics[width=2.4cm]{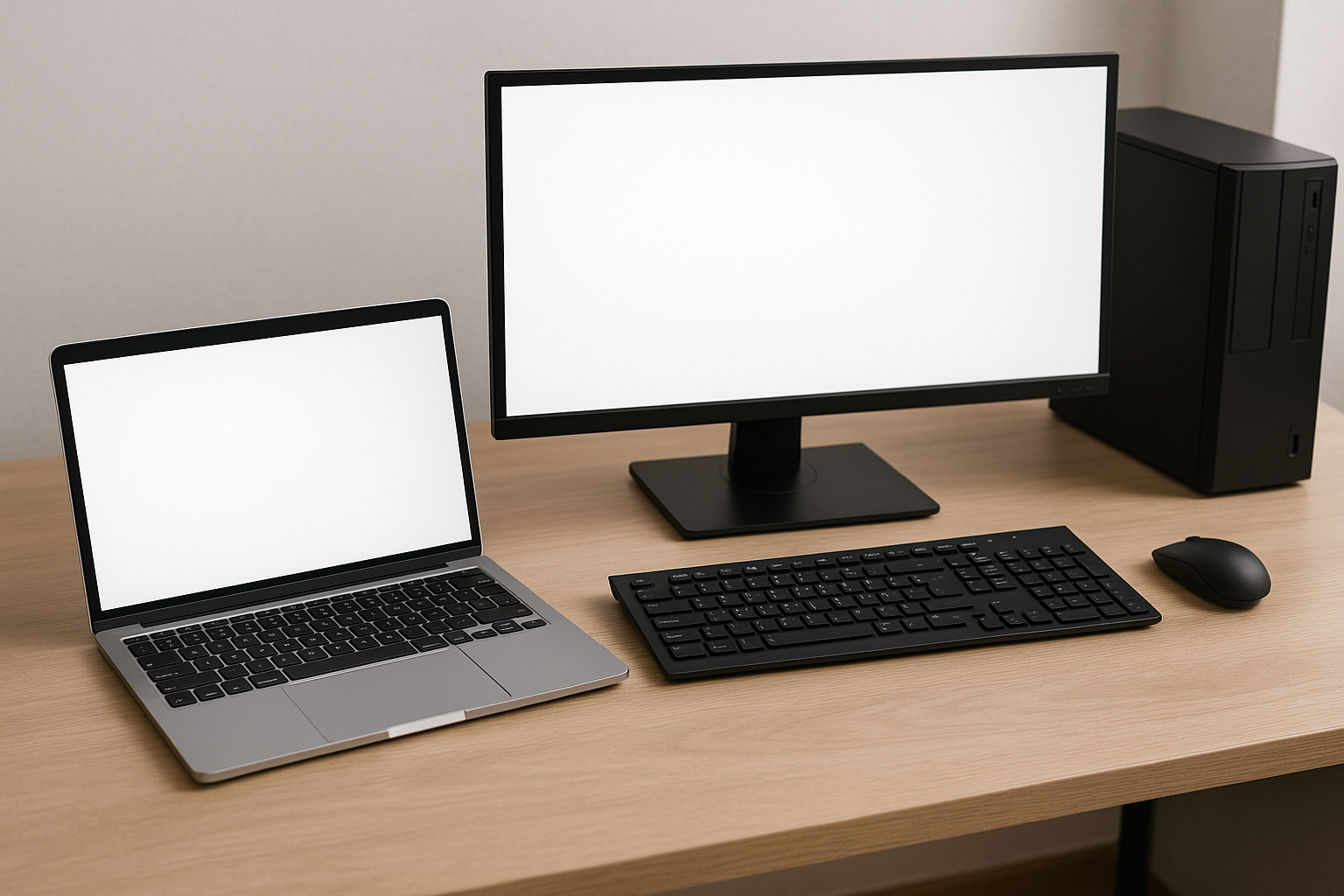}} &
\textcolor{RoyalBlue}{Rank:1, nDCG@10: 1.0} \\ % this is visret with image-1-high, 3 imgs
\\
\\
% \cmidrule{2-6}
% & A residential bathroom with sink, tub, and toilet setting in it.    &
% \multirow{3}*{\includegraphics[width=1.4cm]{figures/qual_examples/mscoco_gt87.jpg}} &
% \textcolor{BrickRed}{Rank:54, nDCG@10: 0.00} &
% \multirow{3}*{\includegraphics[width=1.7cm]{figures/qual_examples/mscoco_gen87.png}} &
% \textcolor{RoyalBlue}{Rank:4, nDCG@10: 0.39} \\ % this is visret with image-1-high, 3 imgs
% \\
% \\
% \\
% \midrule
\bottomrule
\end{tabular}
}
% \vspace{-2mm}
\caption{Additional qualitative results on the three benchmarking datasets.}
\label{tab:additional-qualitative-examples}
\end{table*}

\section{Further Qualitative Examples}
\label{appendix-further-qual-ex}

In \Cref{tab:additional-qualitative-examples}, we present additional qualitative examples on the four evaluation datasets. Consistent with quantitative findings, VisRet reduces the embedding model's workload by expressing the important details in a visualization that is closer to the retrieval target, resulting in substantially better performance in terms of ground-truth rank and nDCG@10.

\begin{figure*}[t!]
    \centering
    \small
    \resizebox{\linewidth}{!}{
    \fbox{\begin{tabular}{ p{\linewidth} }
    \texttt{You are given a query, rephrase the query into a short descriptive phrase that highlights the key part of the entity where the queried feature could be found. DO NOT include the asked feature (shape, color, etc.) but instead include the part of the entity where the feature could be found. Output only the rephrased query.}
    \\
    \\
    \texttt{Examples:}\\
    \\
    \texttt{Original query: What shape are the flowers of bush flax (scientific name: Astelia fragrans)?}
    \\
    \\
    \texttt{Rephrased query: flowers of bush flax}\\
    \\
    \texttt{Original query: Is the any specific color pattern on underside wings of tawny pipit (scientific name: Anthus campestris) displayed during flight, or is it uniformly colored?}
    \\
    \\
    \texttt{Rephrased query: tawny pipit with its underside wings shown}\\
    \\
    \texttt{Original query: \{question\}}\\
    \\
    \texttt{Rephrased query:}
    \end{tabular}}
    }
    % \vspace{-2mm}
    \caption{Prompt for instructing an LLM to generate the T2I generation instruction for Visual-RAG questions.}
    % \vspace{-1mm}
    \label{t2i-ins-prompt-visual-rag}
    % \vspace{-2mm}
\end{figure*}

\begin{figure*}[t!]
    \centering
    \small
    \resizebox{\linewidth}{!}{
    \fbox{\begin{tabular}{ p{\linewidth} }
    \texttt{You are given a query about two entities, as well as an entity of interest. Rephrase the query into a short descriptive phrase that highlights the key part of the entity of interest on which the queried feature could be found. DO NOT include the asked feature (shape, color, etc.) but instead include the entity name + part of the entity where the feature could be found. Output only the rephrased query.}
    \\
    \\
    \texttt{Examples:}\\
    \\
    \texttt{Original query: Are the tongues of grass snake (scientific name: Natrix helvetica) and Chicken Snake (scientific name: Spilotes pullatus) the same color?}
    \\
    \\
    \texttt{Entity of interest: Spilotes pullatus}
    \\
    \\
    \texttt{Rephrased query: Chicken Snake with its tongue shown}\\
    \\
    \texttt{Original query: Which one has a more slender matured legume, common milkpea (scientific name: Galega officinalis) or narrowleaf lupin (scientific name: Lupinus angustifolius)?}
    \\
    \\
    \texttt{Entity of interest: Galega officinalis}
    \\
    \\
    \texttt{Rephrased query: the legume of common milkpea}\\
    \\
    \texttt{Original query: \{question\}}\\
    \\
    \texttt{Entity of interest: \{entity\}}
    \\
    \\
    \texttt{Rephrased query:}
    \end{tabular}}
    }
    % \vspace{-2mm}
    \caption{Prompt for instructing an LLM to generate the T2I generation instruction for Visual-RAG-ME questions.}
    % \vspace{-1mm}
    \label{t2i-ins-prompt-visual-rag-me}
    % \vspace{-2mm}
\end{figure*}

\begin{figure*}[t!]
    \centering
    \small
    \resizebox{\linewidth}{!}{
    \fbox{\begin{tabular}{ p{\linewidth} }
    \texttt{You are given an image retrieval query, rephrase the query to add in a bit detail (no longer than 30 words). The rephrased query should highlight the appearance, posture, actions of the main entity so that it is easier to retrieve the desired image among (1) images of the same entity with different posture and (2) images of different entities with the same posture.}
    \\
    \\
    \texttt{Original query: \{question\}}\\
    \\
    \texttt{Rephrased query:}
    \end{tabular}}
    }
    % \vspace{-2mm}
    \caption{Prompt for instructing an LLM to generate the T2I generation instruction for INQUIRE-Rerank-Hard and COCO-Hard questions.}
    % \vspace{-1mm}
    \label{t2i-ins-prompt-inquire}
    % \vspace{-2mm}
\end{figure*}

\begin{figure*}[t!]
    \centering
    \small
    \resizebox{\linewidth}{!}{
    \fbox{\begin{tabular}{ p{\linewidth} }
    \texttt{Given a question from the user regarding a visual feature of an organism (animal, plant, etc.), answer it using systematic reasoning. You will be provided with one or more images that may contain the key information for answering the question. Your output should consist of two parts.} \\
\\
\texttt{1. Reasoning:} \\
\texttt{- Look at the image carefully. Pick out the feature that can help you correctly answer the question.} \\
\texttt{- If no useful information can be inferred from the image, you should summarize your own knowledge related to the question.} \\
\texttt{- If the image contradicts your own knowledge, you should trust the image.} \\
\texttt{- If the image is blurry, you should summarize your own knowledge related to the question.} \\
\\
\texttt{2. Answer:} \\
\texttt{- Only your conclusion that directly answers the question.} \\
\texttt{- No need to repeat the reasoning.} \\
\\
\texttt{Please always follow the answer format without bolding texts: "\#\#\# Reasoning: \{reasoning\}\textbackslash n\#\#\# Answer: \{your\_answer\}"}
    \end{tabular}}
    }
    % \vspace{-2mm}
    \caption{Prompt for VQA on Visual-RAG.}
    % \vspace{-1mm}
    \label{qa-prompt-visrag}
    % \vspace{-2mm}
\end{figure*}

\begin{figure*}[t!]
    \centering
    \small
    \resizebox{\linewidth}{!}{
    \fbox{\begin{tabular}{ p{\linewidth} }
    \texttt{You are a model that rigorously answers a question that compares a visual feature of two organisms (animal, plant, etc.) using systematic reasoning. You will be provided with one or more images of both organisms that may contain the key information for answering the question. Your output should consist of two parts.} \\
\\
\texttt{1. Reasoning:} \\
\texttt{- Look at the images carefully. Pick out the features that can help you correctly answer the question.} \\
\texttt{- If no useful information can be inferred from the image, you should summarize your own knowledge related to the organism.} \\
\texttt{- If the image contradicts your own knowledge, you should trust the image.} \\
\texttt{- If the image is blurry, you should summarize your own knowledge related to the question.} \\
\texttt{- Then, compare the features of the two organisms and reason through the question step by step.} \\
\texttt{- Finally, conclude your reasoning with a verification step that confirms the correctness of your answer based on the evidence you have gathered.} \\
\\
\texttt{2. Answer:} \\
\texttt{- Only your conclusion that directly answers the question.} \\
\texttt{- No need to repeat the reasoning.} \\
\\
\texttt{Please always follow the answer format without bolding texts: "\#\#\# Reasoning: \{reasoning\}\textbackslash n\#\#\# Answer: \{your\_answer\}"}%

    \end{tabular}}
    }
    % \vspace{-2mm}
    \caption{Prompt for VQA on Visual-RAG-ME.}
    % \vspace{-1mm}
    \label{qa-prompt-visrag-me}
    % \vspace{-2mm}
\end{figure*}

\begin{figure*}[t!]
    \centering
    \vspace{-5mm}
    \small
    \resizebox{\linewidth}{!}{
    \fbox{\begin{tabular}{p{\linewidth}}
\texttt{Please evaluate the answer to a question, score from 0 to 1. The reference answer is provided, and the reference is usually short phrases or a single keyword. If the student answer is containing the keywords or similar expressions (including similar color), without any additional guessed information, it is full correct. If the student answer have missed some important part in the reference answer, please assign partial score. Usually, when there are 2 key features and only 1 is being answered, assign 0.5 score; if there are more than 2 key features, adjust partial score by ratio of correctly answered key feature. The reference answer can be in the form of a Python list, in this case, any one of the list item is correct.}\\
\\
\texttt{If student answer contain irrelevant information not related to question, mark it with "Redundant", but it does not affect score if related part are correct. (e.g. Question: what shape is leave of Sanguinaria canadensis, Student Answer: shape is xxx, color is yyy, this is Redundant answer)}\\
\\
\texttt{If student answer contain features not listed in reference answer, mark it with "Likely Hallucination" and deduct 0.5 score. (e.g., Reference Answer: black and white. Student Answer: black white, with yellow dots, “yellow dots” is not mentioned in reference). The reference answer sometimes contains an add-on enclosed by brackets (), to help verifying hallucinations (e.g.: “shape is xxx (color is yyy)”). Not mentioning add-on information in answer is not considered wrong. Answering "I don't know", "Not enough information" is considered wrong.}\\
\\
\texttt{Format Instructions: Separate the remarks with score using "|", that is, use the syntax of: "Score: \{score\} | Likely Hallucination", "Score: \{score\}", "Score: \{score\} | Likely Hallucination | Redundant". If any explanation on why giving the score is needed, do not start a new line and append after remark with brackets, e.g. "Score: \{score\} | Redundant | (Explanation: abc)".}\\
\\
\texttt{Following are few examples:}\\
\\
\texttt{Question: Is there any specific color marking around the eyes of a semipalmated plover (scientific name: Charadrius semipalmatus)?}\\
\texttt{Reference Answer: black eye-round feather, white stripe above eyes. (sometimes connected to the white forehead)}\\
\\
\texttt{Student Answer: Yes, the bird has a distinctive black line that runs through the eye, which is a key identifying feature.}\\
\texttt{Score: 0 | Likely Hallucination}\\
\\
\texttt{Student Answer: They have a black vertical band in front of the eye, a white band above the eye, and a single black band that wraps partially around the eye, creating a partial "mask" appearance.}\\
\texttt{Score: 1}\\
\\
\texttt{Student Answer: Yes, the semipalmated plover has a distinctive black/dark ring around its eye, surrounded by a bright white ring or patch}\\
\texttt{Score: 0.5 | Likely Hallucination (Explanation: not white ring, but only a line above the eye)}\\
\\
\texttt{Question: What is the typical color of the antennae of Harris's checkerspot butterfly (scientific name: Chlosyne harrisii)?}\\
\texttt{Reference Answer: alternating black and white band, with yellow on the tip}\\
\\
\texttt{Student Answer: The antennae of Harris's checkerspot butterfly are black with orange-tipped clubs.}\\
\texttt{Score: 0.5 (Explanation: not mentioning black and white)}\\
\\
\texttt{Student Answer: The typical color of the antennae of Harris's checkerspot butterfly is black with white spots.}\\
\texttt{Score: 0.5 | Likely Hallucination (Explanation: not white spot but band. Not mentioning the tip)}\\
\\
\texttt{Question: Are the leaves of burro-weed (scientific name: Ambrosia dumosa) usually covered in small hairs?}\\
\texttt{Reference Answer: yes}\\
\\
\texttt{Student Answer: Yes, the leaves of burro-weed (Ambrosia dumosa) are typically covered in small hairs, giving them a grayish or whitish-green appearance.}\\
\texttt{Score: 1 | Redundant}\\
\\
\texttt{Now, score the following question:}\\
\\
\texttt{Question: \{question\}}\\
\texttt{Reference Answer: \{reference\_answer\}}\\
\\
\texttt{Student Answer: \{model\_answer\}}
\end{tabular}}
}
% \vspace{-2mm}
\caption{Prompt for the LLM VQA judge used for Visual-RAG and Visual-RAG-ME.}
\label{eval-prompt}
\end{figure*}

\end{document}